\documentclass[journal]{IEEEtran}

\usepackage{relsize}

\usepackage{placeins}
\usepackage[inline]{enumitem}

\ifCLASSINFOpdf
\else
\fi

\usepackage{amsmath} 
\usepackage{todonotes}
\usepackage{multirow}
\usepackage{array}
\usepackage[nolist]{acronym}
\usepackage{hyperref}

\ifCLASSOPTIONcompsoc
\usepackage[caption=false,font=normalsize,labelfon
t=sf,textfont=sf]{subfig}
\else
\usepackage[caption=false,font=footnotesize]{subfi
g}
\fi

\newcolumntype{P}[1]{>{\centering\arraybackslash}p{#1}}
\newcommand{\Figref}[1]{Fig.~\ref{#1}}
\newcommand{\Secref}[1]{Section~\ref{#1}}
\newcommand{\Subsecref}[1]{Section~\ref{#1}}
\newcommand{\Subsubsecref}[1]{Section~\ref{#1}}
\newcommand{\Eqref}[1]{Eq.~\ref{#1}}
\newcommand{\Tabref}[1]{Table~\ref{#1}}

\begin{document}
\begin{acronym}
  \acro{BIECON}{blind image evaluator based on a convolutional neural network}
  \acro{CNN}{convolutional neural network}
  \acro{CLIVE}{LIVE In the Wild Image Quality Challenge Database}
  \acro{DOG}{difference of Gaussian}
  \acro{FR}{full-reference}
  \acro{FSIM}{feature similarity index}
  \acro{GMSD}{gradient magnitude similarity deviation}
  \acro{HaarPSI}{Haar wavelet-based perceptual similarity index}
  \acro{HVS}{human visual system}
  \acro{IQA}{image quality assessment}
  \acro{IQM}{image quality measure}
  \acro{KROCC}{Kendall rank order coefficient}
  \acro{LCC}{Pearson linear correlation coefficient}
  \acro{NR}{no-reference}
  \acro{RR}{reduced-reference}
  \acro{MAE}{mean absolute error}
  \acro{MOS}{mean opinion score}
  \acro{MSE}{mean square error}
  \acro{QA}{quality assessment}
  \acro{PCA}{principal component analysis}
  \acro{PSNR}{peak signal-to-noise ratio}
  \acro{ReLU}{rectified linear unit}
  \acro{SROCC}{Spearman rank order coefficient}
  \acro{SSIM}{structural similarity index}
  \acro{SVR}{support vector regression}
  \acro{DIQaM-FR}{Deep Image QuAlity Measure for FR IQA}
  \acro{WaDIQaM-FR}{Weighted Average Deep Image QuAlity Measure for FR IQA}
  \acro{DIQaM-NR}{Deep Image QuAlity Measure for NR IQA}
  \acro{WaDIQaM-NR}{Weighted Average Deep Image QuAlity Measure for NR IQA}
\end{acronym}

\title{Deep Neural Networks for No-Reference and Full-Reference Image Quality
Assessment}

\author{Sebastian Bosse$^{\dag}$,
Dominique Maniry$^\dag$,
Klaus-Robert M\"uller, \textit{Member, IEEE},\\
Thomas Wiegand, \textit{Fellow, IEEE},
and Wojciech Samek, \textit{Member, IEEE}
\thanks{$^\dag$SB and DM contributed equally.}

 \thanks{S. Bosse, D. Maniry and W. Samek are with the Department of Video
Coding \& Analytics, Fraunhofer Heinrich Hertz Institute (Fraunhofer HHI), 10587
Berlin, Germany (e-mail: sebastian.bosse@hhi.fraunhofer.de;
\mbox{wojciech.samek@hhi.fraunhofer.de)}.}

\thanks{ K.-R. M\"uller is with the Machine Learning Laboratory, Berlin Institute of Technology, 10587 Berlin, Germany, the Department of Brain and Cognitive
Engineering, Korea University, Anam-dong, Seongbuk-gu, Seoul 136-713, Korea, and also with the Max Planck Institute for Informatics, 66123 Saarbr{\"u}cken, Germany
(e-mail: klaus-robert.mueller@tu-berlin.de).}

\thanks{T. Wiegand is with the Fraunhofer Heinrich Hertz Institute (Fraunhofer HHI), 10587 Berlin, Germany
and with the Media Technology Laboratory, Berlin Institute of Technology, 10587 Berlin, Germany
(e-mail: \mbox{twiegand@ieee.org}).} }

\maketitle

\begin{abstract}
We present a deep neural network-based approach to image quality assessment (IQA). 
The network is trained end-to-end and comprises 10 convolutional layers and 5 pooling layers for
feature extraction, and 2 fully connected layers for regression, which makes it significantly deeper than related IQA models.
Unique features of the proposed architecture are that (1) with
slight adaptations it can be used in a no-reference (NR) as well as in a full-reference (FR)
IQA setting and (2) it allows for joint learning of local quality and local weights, i.e., relative
importance of local quality to the global quality estimate, in an
unified framework.
Our approach
is purely data-driven and does not rely on hand-crafted features or other types of prior domain knowledge about the human visual system or image statistics.
We evaluate the proposed approach on the LIVE, CISQ and TID2013 databases as well as the LIVE In the Wild Image Quality Challenge Database and show superior performance to state-of-the-art NR and FR IQA methods.
Finally, cross-database evaluation shows a high ability to generalize between different databases,
indicating a high robustness of the learned features. 
\end{abstract}


\IEEEpeerreviewmaketitle


%
%

\section{Introduction}
\label{sec:intro}
\IEEEPARstart{D}{igital} video is ubiquitous today in almost every
aspect of life, and applications such as  high definition television,
video chat, or internet video streaming are used for information programs, entertainment and
private communication.
In most applications, digital images are intended to be viewed by
humans. Thus, assessing its perceived quality is essential for most problems in
image communication and processing. In a communication system, for example,
when an image arrives at the ultimate receiver, typically a human viewer, it has passed a
pipeline of  processing  stages, such  as  digitization, compression or
transmission. These  different  stages introduce  distortions into the original
image, possibly visible to human viewers, that may exhibit a 
certain level of  annoyance in  the viewing experience. Hence,
 predicting perceived visual quality computationally is crucial for the 
optimization and evaluation of such a system and its modules.

Generally, \acp{IQM} are classified  depending on the
amount of information available from an original reference image --- if existing at all.
While \ac{FR} approaches have access to the full reference image, 
no information about it is available to \ac{NR} approaches.
Since for \ac{RR} \ac{IQA} only a set
of features extracted from the reference image is available to the algorithm, it lies somewhere in the 
middle of this spectrum.
As no information about the original is exploited, \ac{NR} \ac{IQA} is
a more difficult problem than \ac{FR} \ac{IQA} and potentially the most
challenging problem in \ac{IQA}. \ac{NR} \ac{IQA} has a wide range of applications as in
practice often no reference is available.
However, for many applications, such as the optimization of video coding systems, unconstrained \ac{NR} \ac{IQA} is not a feasible approach
---imagine a video codec that, as one example, reconstructs a noise and blur
free version of the movie \textit{Blair Witch Project} or, as another example,
independently of the input, always reconstructs the same image (but this of
perfect quality).

Perceptually accurate \ac{IQA} relies on computational models  of the \ac{HVS} and/or natural image statistics
and is not an easy task \cite{wang2002image}.
The underlying model employed in an \ac{IQM} allows for conceptual
classification, traditionally in bottom-up and top-down approaches. 
While former approaches are based on a computational system simulating the
\ac{HVS} by modeling its relevant components, latter ones treat the \ac{HVS}
as a black box and track distortion specific deviations in image statistics.
In the \ac{NR} \ac{IQA}
case this is typically done based on a model of general statistical regularities
of natural images \cite{bovik2013automatic}.
Typically, these
statistical image models are used to extract features related to perceived quality that
are input to trainable regression models. 
With the rise of machine learning, recently a third category
of \ac{IQA} emerged, comprising approaches that are purely data-driven, do not rely on any
explicit model and allow for end-to-end optimization of feature extraction and
regression.
Our approach presented in this paper belongs to this new class of
data-driven approaches and employs a deep neural network for \ac{FR}
and \ac{NR} \ac{IQA}.
In order to improve prediction accuracy, \acp{IQM} can be combined with saliency
models. This extension aims at estimating the likelihood of regions 
to be attended to by a human viewer and to consider this likelihood in the
quality prediction.

Until recently, problems in computer vision such as image classification and
object detection were tackled in two steps: (1) designing appropriate features
and (2) designing learning algorithms for regression or classification. Although
the extracted features were input to the learning algorithm, both of these steps were mostly
independent from each other.  
Over the last years, \acp{CNN} have shown to
outperform these traditional approaches. One reason is that they allow
for joint end-to-end learning of features and regression based on the raw input data
without any hand-engineering.
It was further shown that in classification tasks deep \acp{CNN} with more
layers outperform shallow network architectures \cite{Simonyan2015}.

This paper studies the use of a deep CNN with an architecture 
\cite{Simonyan2015} largely inspired by the organization of the primates' visual cortex, comprising 
10 convolutional layers and 5 pooling layers for feature extraction, and 2 fully
connected layers for regression, in a general  \ac{IQA} setting and shows 
that network depth has a significant impact on performance.
We start with addressing the problem of \ac{FR} \ac{IQA} in an end-to-end
optimization framework. For that, we adapt the concept of Siamese networks
known from classification tasks \cite{bromley1993signature,Chopra2005} by
introducing a feature fusion stage in order to allow for a joint regression of the
features extracted from the reference and the distorted image. Different feature fusion
strategies are discussed and evaluated.

As the number of parameters to be trained in deep networks is usually very large,
the training set has to contain enough data samples in order to avoid
overfitting. Since publicly available quality-annotated image databases are
rather small, this renders end-to-end training of a deep network a challenging
task.
We address this problem by artificially augmenting the datasets, i.e., we train
the network on randomly sampled patches of the  quality annotated images. For
that, image patches are assigned quality labels from the corresponding
annotated images. Different to most data-driven  \ac{IQA} approaches, patches input to the
network are not normalized, which  enables the proposed method to also cope with
distortions introduced by luminance and  contrast changes. To this end, global
image quality is derived by pooling local patch qualities simply by averaging
and, for convenience, this method is referred to as \ac{DIQaM-FR}.

However, neither local quality nor relative importance of local qualities is uniformly distributed
over an image. This leads to a high amount of  label noise in the augmented datasets.
 Thus, as a
further contribution of this paper, we assign a patchwise relative weight to account for its influence on the global
quality estimate. This is realized by a simple change to the network 
architecture and adds two fully connected layers running in parallel to the
quality regression layers, combined with a modification of the training
strategy. We refer to this method as  \ac{WaDIQaM-FR}.
This approach allows for a joint optimization of local quality assessment and
pooling from local to global quality, formally within the classical framework of saliency weighted
distortion pooling.  

After establishing our approaches within a \ac{FR} \ac{IQA} context, we
abolish one of the feature extraction paths in the Siamese network. 
This adaptation allows to apply the network within a \ac{NR} context as well. Depending on the spatial
pooling strategy used, we refer to the \ac{NR} \ac{IQA} models as \ac{DIQaM-NR} and \ac{WaDIQaM-NR}.

Interestingly, starting with a \ac{FR} model, our approach facilitates
systematic reduction of the amount of information from the reference image needed for accurate
quality prediction. Thus, it helps closing the gap between \ac{FR} and
\ac{NR} \ac{IQA}. 
We show that this allows for exploring the space of \ac{RR} \ac{IQA} from a given \ac{FR} model without retraining.
In order to facilitate reproducible research, our implementation is made publicly available at
\texttt{\url{https://github.com/dmaniry/deepIQA}}.

The performance of the \ac{IQA} models trained with the proposed methods are
evaluated and compared to state-of-the-art \acp{IQM} on the popular LIVE,
TID2013 and CISQ image quality databases. The models obtained for \ac{NR} \ac{IQA} 
are additionally evaluated on the more recent LIVE In the Wild Image Quality
Challenge Database (that, for convenience, we will refer to as CLIVE).
Since the performance of data-driven approaches largely relies on on the data
used for training we analyze the generalization ability of the proposed methods
in cross-database experiments.

The paper is structured as follows: In \Secref{sec:relatedWork} we give an
overview over state-of-the-art related to the work presented in this
paper.
\Secref{sec:methods} develops and details the proposed methods for deep neural
network-based \ac{FR} and \ac{NR} \ac{IQA} with different patch aggregation
methods. 
In \Secref{sec:results} the presented approaches are evaluated and compared to
related \ac{IQA} methods. Further, weighted average patch aggregation, network
depth, and reference reduction are analyzed.
We conclude the paper with a discussion and an
outlook to future work in \Secref{sec:discussion}.

\section{Related Work}
\label{sec:relatedWork}
\subsection{Full-Reference Image Quality Assessment}
\label{ssec:rel_work_fr}
The most simple and straight-forward image quality metric is 
the \ac{MSE} 
between reference and distorted image. Although being widely used, it does not
correlate well with perceived visual quality \cite{Gir93}. This led to the development of
a whole zoo of  image  quality metrics that strive for a better agreement with
the image quality as perceived by humans \cite{Lin2011}. 

Most popular quality metrics belong to the class of top-down approaches and try
to identify and exploit distortion-related changes in image features in order to
estimate perceived quality. These kinds of approaches can be found in the
\ac{FR}, \ac{RR}, and \ac{NR} domain.
The SSIM \cite{Wang2004}  is probably the most prominent example of
these approaches. It considers the sensitivity of the \ac{HVS} to
structural information by  pooling luminance similarity (comparing local mean luminance),
contrast similarity (comparing local variances) and structural similarity 
(measured as local covariance). The SSIM was not only extended for multiple
scales to the MS-SSIM \cite{Wang2003}, but  the framework of pooling
complementary features similarity maps served as inspiration for other \ac{FR} \ac{IQM}s
employing different features, such as the FSIM \cite{Zhang2011}, the GMSD \cite{Xue2014}, the SR-SIM
\cite{ZhLi2012} or HaarPSI \cite{Reisenhofer2016}.
DeepSim \cite{gao2017deepsim} extracts feature maps for similarity computation from the different
layers of a deep \ac{CNN}  pre-trained for recognition, showing that features learned for image
recognition are also meaningful in the context of perceived quality.
The \ac{DOG}-SSIM \cite{Pei2015} belongs somewhat to the top-down as well
as to the bottom-up domain, as it mimics the frequency bands of the contrast
sensitivity function using a \ac{DOG}-based channel decomposition. Channels are
then input to SSIM in order to calculate channel-wise quality
values that are pooled by a trained regression model to an overall quality estimate.
The MAD \cite{LaCh2010} distinguishes between supra- and near-threshold distortions to account for
different domains of human quality perception.

Combining several hand-crafted \acp{IQM} can have better performance
than any single \ac{IQM} in the set \cite{Lukin2015}.

Although feature learning \cite{gao2017deepsim} and regression \cite{Pei2015,Lukin2015}
have been employed in \ac{FR} \ac{IQA}, to our best knowledge, so far no end-to-end
trained method was used for this task. 

\subsection{No-Reference Image Quality Assessment}
A typical approach to
\ac{NR} \ac{IQA} is to model  statistics of natural  images  and regress parametric deviations from this model
to perceived image  degradations.
As these  parameters and its deviations may depend on the distortion type, the 
DIIVINE framework \cite{Moorthy2011}  identifies the distortion type affecting an  image in a
first step and uses a distortion-specific regression scheme  to estimate the
perceived quality in a second step. The statistical features are  calculated
based on an oriented subband decomposition.
BLIINDS-II \cite{Saad2012} uses a generalized Gaussian density function to model
block  DCT coefficients of images.
BRISQUE \cite{Mittal2012} proposes a \ac{NR} \ac{IQA} approach that utilizes an
asymmetric generalized Gaussian distribution to model images in the spatial
domain. The modeled image features here are differences of  spatially
neighbored,  mean subtracted and contrast normalized image samples.
NIQE \cite{Mittal2013} extracts features based on a multivariate Gaussian
model and relates them to perceived quality in an unsupervised manner.
In order to cope with more complex and authentic distortion types FRIQUEE
\cite{Ghadiyaram2015,ghadiyaram2016massive} employs a deep belief network of 4
layers trained to classify bins of 10 different distortion ranges. Input to the
network is a set of handcrafted feature maps and the feature representation on the last
hidden   layer is extracted to be input to \ac{SVR} for quality prediction.

CORNIA \cite{Ye2012a} is one of the first purely data-driven \ac{NR} \ac{IQA}
methods combining feature and regression training. Here, a codebook is
constructed by k-means clustering of luminance and contrast normalized image patches.
Soft-encoded distances between  visual codewords and  patches extracted
from distorted images are used as features that are pooled and regressed using \ac{SVR} for
estimating image quality. This approach is refined to the semantic  obviousness metric (SOM)
\cite{Zhang2015}, where object-like regions are detected and the patches extracted from these regions are
input to CORNIA. Similarly to CORNIA, QAF  \cite{zhang2014training} constructs a
codebook using sparse filter learning based on image log-Gabor responses. As
log-Gabor responses are often considered a  low level model of the HVS,
conceptually, QAF also belongs to the bottom-up  domain. 

Motivated by the recent
success of \acp{CNN} for classification and  detection tasks and the notion that
the connectivity patterns in these networks  resemble  those of the primate visual cortex,
\cite{Kang2014} proposes a shallow \ac{CNN}  consisting of  1 convolutional
layer, 1 pooling layer and 2 fully-connected  layers, that  combines feature
extraction and regression. Quality is estimated on contrast normalized image
patches and patchwise quality is pooled to imagewise quality by averaging.
BIECON \cite{kim2017fully} proposes an interesting approach for data augmentation and tackles
\ac{CNN}-based \ac{NR} \ac{IQA} in 2 steps:
First, a local quality is estimated based on normalized image patches employing a
\ac{CNN} of 2 convolutional, 2 pooling and 5 fully-connected layers. This
network is trained to replicate a conventional \ac{FR} \ac{IQM} such as
SSIM or GMSD within a \ac{NR} framework.
Second,  mean values and the standard deviations of the extracted patchwise features are regressed to an imagewise quality estimate
employing a perceptron with one hidden layer.
Preliminary results on
the application of deeper neural networks, trained end-to-end, for \ac{IQA} have been presented in
\cite{bosse2016icipDeep,bosse2016pcs} and are extended and further evaluated in this paper. 

\subsection{Salience and Attention for \ac{IQA}}
\label{ssec:related_saliency}
Not every region in an image receives the same amount of attention by
viewers and generally distortions in regions that attract viewers' attention are
assumed to be more disturbing than in other regions. This led to the idea to
combine models of visual saliency with \acp{IQM} by weighting the local quality
$y_i$ of a region $i$ with the corresponding local saliency $w_i$ to the overall
image quality $Q$ with

\begin{equation}
  Q =  \frac{\sum_i w_iy_i}{\sum_i w_i}
  \label{eq:saliency_weighted_quality}
\end{equation}

Various models of saliency have been proposed and combined with different
\acp{IQM} to improve prediction performance \cite{Zhang2016}. 
The Visual Saliency-Induced Index (VSI) \cite{Zhang2014} takes the local saliency from reference and distorted image as features maps in a
similarity formalism (\Subsecref{ssec:rel_work_fr})
and combines it with similarity from local gradients and local chrominance. The combined
similarity maps are then spatially pooled by the local maximal
saliency of reference and distorted image.

Saliency has usually been extracted from the reference image and, as for VSI, in some cases
additionally from the distorted image.
Employing saliency estimated from the reference image shifts \ac{NR} \ac{IQA} to the \ac{RR} domain.
So far, saliency models incorporated in \acp{IQM} were mostly  developed to predict
viewers' attention to specific regions of an image, but not to predict perceived quality explicitly
in a joint optimization approach.


\section{Deep Neural Networks for Image Quality Assessment}
\label{sec:methods}

\subsection{Neural network-based \ac{FR} \ac{IQA}}
\label{ssec:frIqa}
\begin{figure*}[ht]
\centering
\includegraphics[width=\textwidth]{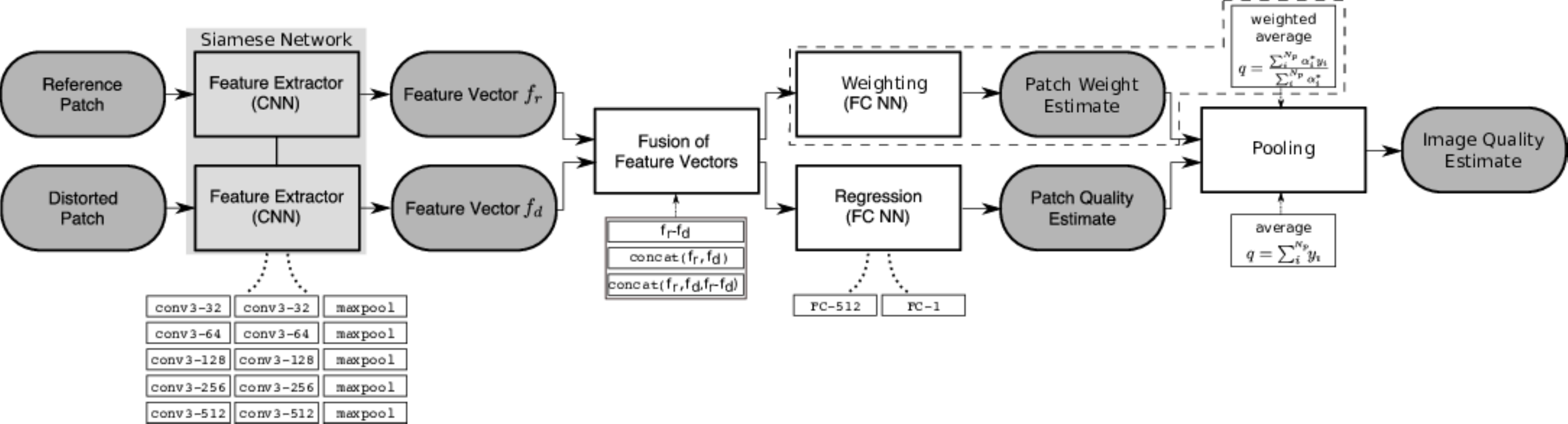}
\caption{Deep neural network model 
for \ac{FR} \ac{IQA}. Features are extracted from the distorted patch and the
reference patch by a \ac{CNN} and fused as difference, concatenation or concatenation
supplementary with the difference vector. 
The fused feature vector is regressed to a patchwise quality estimate.
The dashed-boxed branch of the network indicates an optional regression of the feature vector to a patchwise weight estimate that allows for
pooling by weighted average patch aggregation.} 
\label{fig:fr_network}
\end{figure*}

Siamese networks have been used to learn similarity relations between two
inputs. For this, the inputs are processed in parallel by two networks sharing
their synaptic connection weights. This approach has been used for signature
\cite{bromley1993signature} and face verification \cite{Chopra2005} tasks, where
the inputs are binarily classified as being of the same category or not. 
For \ac{FR} \ac{IQA} we employ a Siamese network for feature extraction. In
order to use the extracted features for the regression problem of \ac{IQA},
feature extraction is followed by a feature fusion step. The fused features are
input to the regression part of the network. 
The architecture of the proposed network is sketched in  \Figref{fig:fr_network}
and will be further detailed in the following.

Motivated by its superior performance in the 2014 ILSRVC classification
challenge \cite{ILSVRC15} and its successful adaptation for various computer vision tasks
\cite{Girshick_2015_ICCV,Long_2015_CVPR}, VGGnet \cite{Simonyan2015} was chosen
as a basis for the proposed networks.
While still a straight-forward, but deep \ac{CNN} architecture,
VGGnet was the first neural network to employ cascaded convolutions kernels small as
$3\times 3$. The input of the VGG network are images of the size $224\times 224$ pixels.
To adjust the network for smaller input sizes such as $32\times 32$ pixel-sizes patches, we extend the network by
3 layers (conv3-32, conv3-32, maxpool) plugged in front of the original architecture.
Our proposed VGGnet-inspired \ac{CNN} comprises 14 weight layers that are 
organized in a feature extraction module and a regression module. 
The features are extracted in a series of  conv3-32, conv3-32, maxpool,
conv3-64, conv3-64, maxpool, conv3-128, conv3-128,  maxpool, conv3-256,
conv3-256, maxpool, conv3-512, conv3-512, maxpool
layers\footnote{\label{fn:notation}Notation is borrowed from \cite{Simonyan2015}
where conv$\langle$receptive field size$\rangle$-$\langle$number of
channels$\rangle$ denotes a convolutional layer and FC$\langle$number of 
channels$\rangle$ a fully-connected layer}. 
The fused features  (see \Subsecref{ssec:feature_pooling}) are regressed  by a
sequence of one FC-512 and one FC-1 layer.
This results  in about 5.2 million trainable network parameters.
All convolutional layers apply $3\times 3$ pixel-size convolution kernels and
are activated through a \ac{ReLU} activation function
$g=\max(0,\sum_i w_ia_i)$, where $g, w_i$ and $a_i$ denote the output, the
weight and the input of the ReLU, respectively  \cite{Nair2010}.
In order to obtain an output of the same size as the input, convolutions are
applied with zero-padding. All  max-pool layers have $2\times 2$ pixel-sized
kernels. 
Dropout regularization with a ratio of 0.5 is applied to the fully-connected
layers in order to prevent overfitting \cite{Srivastava2014}.

For our \ac{IQA} approach, images are subdivided into $32\times 32$ sized
patches that are input to the neural network. Local patchwise qualities are
pooled into a global imagewise quality estimate by simple 
or weighted average patch aggregation.
The choice of strategy for spatial
pooling affects the training of the network and will be explained in more detail
in \Subsecref{ssec:spatial_pooling}. 

For convenience we refer to the resulting models as \acf{DIQaM-FR}
and
\acf{WaDIQaM-FR}.

\subsection{Feature Fusion}
\label{ssec:feature_pooling}
In order to serve as input to the regression part of the network, the
extracted feature vectors $f_r$ and $f_d$ are combined in a feature fusion step. 
In the \ac{FR} \ac{IQA} framework, concatenating $f_r$ and $f_d$ to
$\textrm{\texttt{concat}}(f_r,f_d)$ without any further modifications is
the simplest way of feature fusion. 
$f_r$ and $f_d$ are of identical structure,
which renders the difference $f_r-f_d$ to be a meaningful representation for
distance in feature space. 
Although the regression module should be able to learn $f_r-f_d$ by itself,
the explicit formulation might ease the actual regression task. This allows for
two other simple feature fusion strategies, namely the difference $f_r-f_d$, and 
concatenation to $\textrm{\texttt{concat}}(f_r,f_d,f_r-f_d)$.

\subsection{Spatial Pooling}
\label{ssec:spatial_pooling}
\subsubsection{Pooling by Simple Averaging}
\label{sssec:avgAgg}

The simplest way to pool locally estimated visual qualities $y_i$ to a global
imagewise quality estimate $\hat{q}$ is to assume identical relative importance of
every image region, or, more specifically, of every image patch $i$ as 

\begin{equation}
\hat{q}=\frac{1}{N_p}\sum^{N_p}_{i} y_i,
\end{equation}

where $N_p$ denotes the number of patches sampled from the image.
For regression tasks, commonly the \ac{MSE} is used as minimization criterion.
However, as simple average quality pooling implicitly assigns the locally
perceived quality to be identical to globally perceived quality $q_t$ this
approach introduces a certain degree of label noise into the training data.
Optimization with respect to \ac{MAE} puts less emphasis on outliers and reduces
their influence. As our \ac{IQA} problem is a regression task, we choose
\ac{MAE} as a less outlier sensitive alternative to \ac{MSE}.
The loss function to be minimized is then
\begin{align}
\begin{split}
E_{simple} 
              =& \frac{1}{N_p}\sum^{N_p}_{i}|y_i-q_t|.
\end{split}
\label{eq:patchwiseLoss}
\end{align}

In principle, the number of patches $N_p$ can be chosen arbitrarily.  A
complete set of all non-overlapping patches would ensure all pixels of the image to be
considered and, given the same trained CNN model, be mapped to reproducible
scores.

\subsubsection{Pooling by weighted average patch aggregation}
\label{ssec:wAvgPatchAgg}
As already shortly discussed in \Subsecref{ssec:related_saliency}, 
quality perceived in a local region of an image is not necessarily reflected by the
imagewise globally perceived quality and vice-versa, e.g. due to spatially
non-uniformly distributed distortion, summation or saliency effects or combinations of these influencing factors.
In the above described pooling-by-average approach this is  accounted for only very roughly 
by the choice of a less outlier-sensitive loss function. 
However, spatial pooling by averaging local quality estimates does not consider the
effect of spatially varying perceptual relevance of local quality.

We address the spatial variance of relative image quality by integrating a second branch into the
regression module of the network that runs in parallel to the patchwise quality regression branch
(see \Figref{fig:fr_network}) and shares the same dimensionality.
This branch outputs an $\alpha_i$ for a patch $i$. 
By activating  $\alpha_i$ through
a \ac{ReLU} and adding a small stability term $\epsilon$
\begin{equation}
\alpha_i^*=\max(0,\alpha_i)+\epsilon
\end{equation}
it is guarantied to be $\alpha_i^*>0$ and can be 
used to weight the estimated quality $y_i$ of the respective patch $i$.

With the normalized weights
 \begin{equation}
 p_i = \frac{\alpha^*_i}{\sum^{N_p}_{j}\alpha^*_j}.
 \label{eq:normalizedWeights}
\end{equation}
the global image quality estimate $\hat{q}$ can be calculated as
\begin{align}
\hat{q} &= \sum^{N_p}_{i}p_iy_i
        \ =\  \frac{\sum^{N_p}_i \alpha^*_iy_i}{\sum^{N_p}_i \alpha^*_i} \label{eq:weighted_q}.
\end{align}
As in \Eqref{eq:patchwiseLoss},  the number of patches $N_p$ can be set
arbitrarily.
Comparing \Eqref{eq:weighted_q} to \Eqref{eq:saliency_weighted_quality} shows that the proposed
pooling method implements a weighting technique formally equivalent to 
the framework of linear saliency weighting as described in \Subsecref{ssec:related_saliency}.

For joint end-to-end training the loss function to be minimized is then
\begin{align}
E_{weighted} =& |\hat{q}-q_t|.       
\end{align}

\subsection{Network Adaptations for \ac{NR} \ac{IQA}}
\label{ssec:nrIqa}

\begin{figure*}[ht]
\centering
\includegraphics[width=\textwidth]{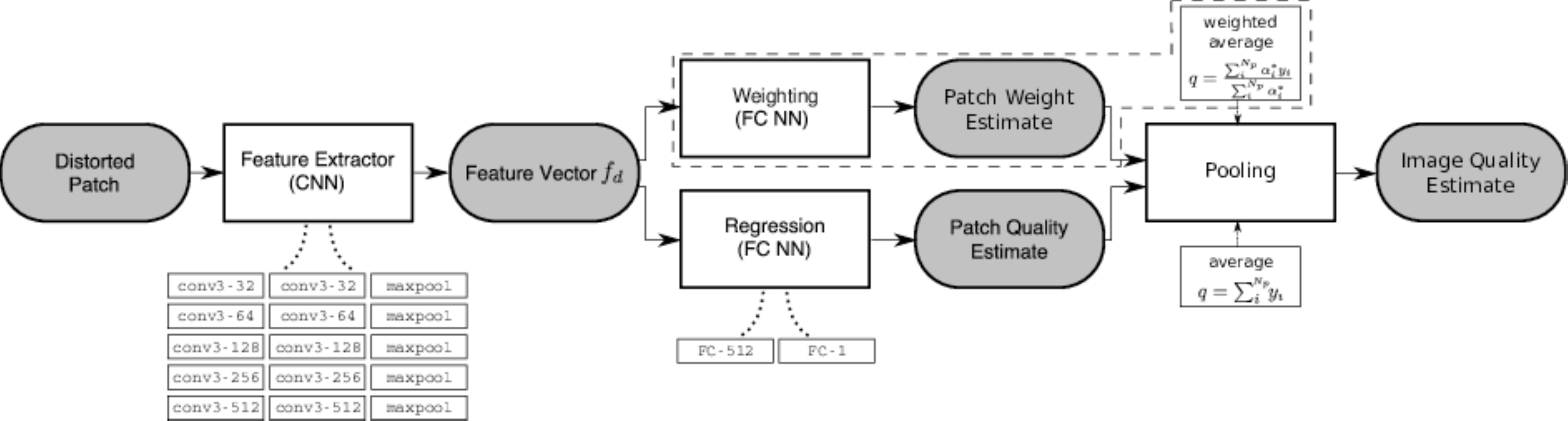}
\caption{Deep neural network for
\ac{NR} \ac{IQA}.
Features are extracted from the distorted patch by a \ac{CNN}. The feature
vector $f_d$ is regressed to a patchwise quality estimate. Patchwise estimates
are aggregated to a global image quality estimate. The dashed-boxed branch of the
network indicates an optional regression of the feature vector to a patchwise weight estimate that allows for
pooling by weighted average patch aggregation.}
\label{fig:nr_network}
\end{figure*}
Abolishing the branch that extracts features from the
reference patch from the Siamese network is a straight forward approach to use
the proposed deep network in a \ac{NR} \ac{IQA} context. As no features
from the reference patch are available anymore, no feature pooling is necessary.
However, both spatial pooling methods detailed in \Subsecref{ssec:spatial_pooling} are
applicable for \ac{NR} \ac{IQA} as well. The resulting approaches are referred to as 
\acf{DIQaM-NR}
and   \acf{WaDIQaM-NR}. This amounts to the same loss functions as for
the \ac{FR} \ac{IQA} case.
The resulting architecture of the neural network adapted for \ac{NR} \ac{IQA} is
illustrated in \Figref{fig:nr_network}.

\subsection{Training}
\label{ssec:training}
The proposed networks are trained iteratively by backpropagation
\cite{LeCun1998, lecun2012efficient} over a number of epochs, where one epoch is defined as the
period during which each sample from the training set has been used once.
In each epoch the training set  is divided into mini-batches for batchwise
optimization.
Although it is possible  to treat each image patch as a separate sample in the
case of simple average pooling, for weighting average pooling image patches of
the same image can not be distributed over different mini-batches, as their
output is combined for the calculation of the normalized weights in the last layer. 
In order to train all methods as similar as possible, each mini-batch contains 4
images, each represented by 32 randomly sampled image patches which leads to the
effective batch size of 128 patches.
The backpropagated error is the average loss over the images in a mini-batch.
For training the \ac{FR} \ac{IQA} networks, 
the respective reference patches are included in the mini-batch.
Patches are randomly sampled every epoch to ensure that as many different image
patches as possible are used in training. 

The learning rate for the batchwise optimization is controlled per parameter adaptively using the ADAM
method \cite{kingma2014adam} based on the variance of the gradient.
Parameters of ADAM are chosen as recommended in \cite{kingma2014adam} as 
$\beta_1=0.9$, $\beta_2=0.999$, $\epsilon=10^{-8}$ and $\alpha=10^{-4}$.
The mean loss over all images during validation is
computed in evaluation mode (i.e. dropout is replaced with scaling) after each 
epoch. The 32 random patches for each validation image are only sampled once at 
the beginning of training in order to avoid noise in the validation loss. The final model used in
evaluation is the one with the best validation loss. This amounts to early stopping \cite{prechelt2012early}, a regularization 
to prevent overfitting.
 
Note that the two regression branches estimating patch weight and patch quality
do not have identical weights, as the update of the network weights
is calculated based on gradients with respect to
different parameters.

\section{Experiments and Results}
\label{sec:results}
\subsection{Datasets}
Experiments are performed on the LIVE \cite{Sheikh2006statistical}, TID2013
\cite{Ponomarenko2013} and  CSIQ \cite{larson2009consumer} image  quality
databases, the \ac{NR} \ac{IQA} approach is also evaluated on the  \acl{CLIVE}
\cite{Ghadiyaram2015,ghadiyaram2016massive}. 

The LIVE \cite{Sheikh2006statistical} database comprises 779 quality annotated
images based on 29 source reference images that are subject to 5 different types
of distortions at different distortion levels. Distortion types are JP2K
compression, JPEG compression, additive white Gaussian noise, Gaussian blur and
a simulated fast fading Rayleigh  channel. Quality ratings were collected using
a single-stimulus methodology,   
scores from different test sessions were aligned. Resulting DMOS quality
ratings lie in the range of $[0,\ 100]$, where a lower
score indicates better visual image quality.

The TID2013 image quality database \cite{Ponomarenko2013} is an extension of the
earlier published TID2008 image quality database \cite{Ponomarenko2009} 
containing 3000 quality annotated images based on 25 source reference images
distorted by 24 different distortion types at 5 distortion levels each. The
distortion types cover a wide range from  simple Gaussian noise or blur over
compression distortions such as JPEG to more  exotic distortion types such as 
non-eccentricity pattern noise. This makes the  TID2013 a more challenging 
database for the evaluation of \acp{IQM}. The rating  procedure differs from the
one used for the construction of LIVE, as it employed a competition-like double stimulus procedure. 
The obtained \ac{MOS} values lie in the
range $[0,\ 9]$, where  larger \ac{MOS} indicate better visual quality. 

The CISQ image quality database contains 866 quality annotated images. 30
reference images are  distorted by JPEG compression, JP2K compression,
Gaussian blur, Gaussian white noise, Gaussian pink noise or contrast change. For
quality assessment, 
subjects were asked to position distorted images horizontally on a monitor according to its visual
quality.
 After alignment and
normalization resulting DMOS values span the range  $[0,\ 1]$, where a lower value
indicates better visual quality. 

The \ac{CLIVE} \cite{Ghadiyaram2015,ghadiyaram2016massive} comprises
1162 images taken under real life conditions with a large variety of objects and scenes captured
under varying luminance conditions using different cameras. In that sense the images are  authentically distorted with
impairments being the result of a  mixture of different distortions, such as
over- or underexposure, blur, grain,  or compression.
As such, no undistorted reference images are available.
Quality annotations were obtained in the form of \ac{MOS}  in a crowdsourced
online study. \ac{MOS} values lie in the range $[0,\ 100]$, a higher value
indicates higher quality.

\subsection{Experimental Setup}
\label{ssec:expSetup}
For evaluation, the networks are trained either on LIVE or TID2013
database.
For cross-validation, databases are randomly split by reference image. This
guarantees that no distorted or undistorted version of an image used in
testing or validation has been seen by the network during training.
For LIVE, the training set is based on 17 reference images, 
validation and test set on 6 reference images each.
TID2013 is split analogously in 15 training, 5 validation and 5 test images. 
CLIVE does not contain versions of different quality levels of the same image,
therefore splitting in sets can be done straight
forward by distorted image. Training set size for CLIVE is 698 images, validation and test set size
232 images each.
Results  reported are based on 10 random splits.
Models are trained for 3000 epochs. Even though some models converge after less than 1000 epochs
a high number is used to ensure convergence for all models. 
During training the network has seen
${\raise.17ex\hbox{$\scriptstyle\sim$}} 48 \textrm{M}$ patches 
in the case of LIVE, ${\raise.17ex\hbox{$\scriptstyle\sim$}} 178 \textrm{M}$ patches in the case of
TID2013, and  ${\raise.17ex\hbox{$\scriptstyle\sim$}} 67 \textrm{M}$ in the case of CLIVE.

To assess the generalization ability of the proposed methods the CSIQ image
database is used for cross-dataset evaluating the models trained either on LIVE
or on TID2013.  For training the model on LIVE, the dataset is split into 23 reference
images for training and 6 for validation; analogously, for training the model on TID2013, 
the dataset is split into 20 training images and 5 validation images.
LIVE and TID2013 have a lot of reference images in common, thus, tests between
these two are unsuitable for evaluating generalization for unseen images.
For cross-distortion evaluation models trained on LIVE are tested on  TID2013
in order to determine how well a model deals with distortions that have not been
seen during training in order to evaluate whether a method is
truly non-distortion or just many-distortion specific.

Note that different to many evaluations reported in the literature, we use
the full TID2013 database and do not ignore any specific distortion type.
To make errors and gradients comparable for different databases, the
\ac{MOS} values of TID2013 and CLIVE and the DMOS values of CSIQ have been linearly mapped to the same  range as
the DMOS values in LIVE. Note that by this mapping high values of $y_i$ represent high local distortion.
For evaluation, prediction accuracy is quantified by \ac{LCC}, prediction
monotonicity is measured by \ac{SROCC}. For both correlation metrics a value
close to 1 indicates high performance of a
specific quality measure.

\subsection{Performance Evaluation}
Evaluations presented in this subsection are based on image quality estimation considering $N_P=32$ patches. Other values of $N_p$  will be discussed in
\Subsubsecref{ssec:convergence_evaluation}.
Performances of the \ac{FR} \ac{IQA} models are reported for features fused by 
$\textrm{\texttt{concat}}(f_r,f_d,f_r-f_d)$; the influence of the different
feature fusion schemes are examined in \Subsubsecref{ssec:feature_fusion}.

\label{sssec:fr_singleDataset_results}
\begin{table}[ht]
\caption{Performance Comparison on LIVE and TID2013 Databases}
\begin{tabular}{  c l | c | c | c | c }
     &&\multicolumn{2}{c|}{LIVE} & \multicolumn{2}{c}{TID2013}\\
     \cline{3-6}
    &IQM  & LCC        & SROCC  & LCC   &  SROCC\\
    \hline
  \multirow{7}{*}{\rotatebox[origin=c]{90}{Full-Reference}}&PSNR       & 0.872    
  & 0.876 & 0.675      & 0.687 \\
  &SSIM \cite{Wang2004}        & 0.945      & 0.948   &
  0.790      & 0.742 \\
  &FSIM$_C$ \cite{Zhang2011}        & 0.960      & 0.963 &
  0.877      & 0.851\\
  &GMSD \cite{Xue2014}  & 0.956 & 0.958 &-&- \\
  &DOG-SSIM \cite{Pei2015} & 0.963 & 0.961  & 0.919 &
  0.907\\
  &DeepSim \cite{gao2017deepsim}&0.968&\bf{0.974}&0.872&0.846\\
  &DIQaM-FR (proposed) & 0.977 & 0.966 & 0.880 & 0.859  \\
  &WaDIQaM-FR  (proposed) & \bf{0.980}      & 0.970&
  \bf{0.946}      & \bf{0.940} \\

    \hline
\multirow{11}{*}{\rotatebox[origin=c]{90}{No-Reference}}
&BLIINDS-II\cite{Saad2012}& 0.916 & 0.912 & 0.628 &0.536\\
&DIIVINE  \cite{Moorthy2011} & 0.923  & 0.925  & 0.654&  0.549        \\
&BRISQUE  \cite{Mittal2012} & 0.942  & 0.939 & 0.651 & 0.573          \\
&NIQE     \cite{Mittal2013} & 0.915  & 0.914 & 0.426 & 0.317           \\
&BIECON   \cite{kim2017fully} & 0.962  & 0.961 &    -  & -       \\
& FRIQUEE \cite{Ghadiyaram2015} & 0.930   & 0.950  &   - & -         \\
&CORNIA   \cite{Ye2012a} & 0.935  & 0.942  & 0.613 &0.549         \\
&CNN      \cite{Kang2014} & 0.956  & 0.956 & -& -            \\
&SOM      \cite{Zhang2015} & 0.962  & \bf{0.964} & -& -             \\
&DIQaM-NR (proposed)  & \bf{0.972}  & 0.960  & \bf{0.855} &\bf{0.835}\\
&WaDIQaM-NR (proposed)  & 0.963  & 0.954 &0.787&0.761           \\
    \end{tabular}
\label{tab:performance_comparison}
\end{table}

\begin{table}[htb!]
\centering
\caption{Performance Comparison for Different Subsets of TID2013}
    \begin{tabular}{  l | P{0.5cm} | P{0.6cm} | P{0.61cm} |
    P{0.6cm} | P{0.5cm} | P{0.45cm} }
    
             & Noise & Actual  & Simple  &  Exotic & New & Color \\
    \hline
  PSNR       & 0.822 & 0.825 & 0.913 & 0.597 & 0.618 & 0.535 \\
  SSIM \cite{Wang2004}        & 0.757 & 0.788 & 0.837 & 0.632 & 0.579 & 0.505 \\
  FSIM$_C$ \cite{Zhang2011}  & 0.902  & 0.915 & 0.947 & 0.841 & 0.788 & 0.775 \\
  DOG-SSIM \cite{Pei2015}      & 0.922  & 0.933  & 0.959 & 0.889 & 0.908 & 0.911\\
  DIQaM-NR   & 0.938 & 0.923 & 0.885 & 0.771 & 0.911 & 0.899 \\
  WaDIQaM-NR   & \bf{0.969}  & \bf{0.970}  & \bf{0.971} & \bf{0.925} & \bf{0.941} & \bf{0.934}\\
    
    \end{tabular}
    \label{tab:perfomance_tid2013_subsets}
\end{table}

\subsubsection{Full-Reference Image Quality Assessment}
The upper part of \Tabref{tab:performance_comparison} summarizes the performance
of the proposed \ac{FR} \ac{IQA} models in comparison to other state-of-the-art
methods on the full LIVE and full TID2013 database in terms of \ac{LCC} and \ac{SROCC}. 
With any of the two presented spatial pooling methods, the proposed
approach obtains superior performance to state-of-the-art on LIVE, except for DeepSim evaluated by
\ac{SROCC}.
On TID2013 DIQaM-FR performs better than most evaluated 
state-of-the-art methods, but is outperformed by
DOG-SSIM\footnote{\label{fn:tid2013}Unfortunately, for many state-of-the-art \ac{FR} and \ac{NR}
\ac{IQA} methods no results are reported on TID2013.}.
Here, employing weighted average patch aggregation clearly improves the
performance and WaDIQaM-FR performs better than any other evaluated \ac{IQM}.
This effect can be observed as well in \Tabref{tab:perfomance_tid2013_subsets}
for the groups of different distortion of
TID2013 defined in \cite{Ponomarenko2013}. While DIQaM-FR performs comparable to the state-of-the-art
method, on some groups better, on some worse, WaDIQaM-FR shows superior performance for all grouped
distortion types.
 
\subsubsection{No-Reference Image Quality Assessment}
\label{sssec:nr_iqa_single_database_performance}
\begin{table}
\centering
\caption{Performance Evaluation for \ac{NR} \ac{IQA} on \ac{CLIVE}}
\begin{tabular}{l|c|c}
IQM &  PLCC & SROCC \\
\hline
FRIQUEE \cite{Ghadiyaram2015} & \bf{0.706} & \bf{0.682} \\
BRISQUE \cite{Mittal2012}& 0.610 & 0.602 \\
DIIVINE \cite{Moorthy2011}& 0.5577& 0.5094 \\
BLIINDS-II \cite{Saad2012} & 0.4496 & 0.4049 \\
NIQE \cite{Mittal2013} & 0.4776 & 0.4210 \\
DIQaM-NR (proposed) & 0.601 &0.606 \\
WaDIQaM-NR (proposed) & 0.680 & 0.671 \\
\end{tabular}
\label{tab:performanc_comparison_clive}

\end{table}

\begin{table}
\centering
\caption{PLCC on selected distortion of TID2013}
\begin{tabular}{l|c|c|c|c}
&            GB     &JEPG   &J2K    &LBDDI \\
\hline  
DIQaM-FR &  0.884 & 0.965 & 0.900 & 0.634  \\
WaDIQaM-FR  &  \bf{0.963} & \bf{0.978} & \bf{0.975} & \bf{0.683}  \\
\hline
DIQaM-NR  & \bf{0.872}  & \bf{0.946}  & \bf{0.872}  & 0.479 \\   
WaDIQaM-NR   & 0.618  & 0.726  & 0.816  & \bf{0.664}  \\
\end{tabular}
\label{tab:performance_by_dist_type_tid2013}
\end{table}

Performances of the \ac{NR} \acp{IQM} are compared to other state-of-the-art
 \ac{NR} \ac{IQA} methods in the lower part of \Tabref{tab:performance_comparison}. 
 The proposed model employing simple average pooling (DIQaM-NR) performs best in terms of LCC among
 all methods evaluated, and in terms of SROCC performs slightly worse than SOM. Evaluated on
 TID2013, DIQaM-NR performs superior among all other methods in terms of
 LCC and SRCC\footnotemark[2]. Although no results were reported for BIECON \cite{kim2017fully}
on the TID2013 dataset, this method achieves a relatively high SROCC=0.923 on the older TID2008 database when 5 distortion types (non-eccentricity pattern noise, local block-wise distortions, mean shift, contrast change) are removed from the analysis. Future investigations will show how BIECON performs on the challenging TID2013 database with all distortions included. In contrast to our \ac{FR} \ac{IQA} models for \ac{NR} \ac{IQA} the weighted
 average patch aggregation pooling decreases the prediction performance.

A comparison of performances for the \ac{CLIVE} database is shown in
\Tabref{tab:performanc_comparison_clive}. Quality assessment on CLIVE is much more difficult than on
LIVE or TID2013, thus performances of all methods evaluated are much worse than for the legacy
databases.
WaDIQaM-NR shows prediction performance superior to most other models, but is clearly outperformed by FRIQUEE.
Interestingly and contrasting to the results on LIVE and
TID2013, on CLIVE WaDIQaM-NR performs clearly better than DIQaM-NR.

In order to analyze these apparent contradictory results a little deeper,
\Tabref{tab:performance_by_dist_type_tid2013} shows the performance of (Wa)DIQaM-FR and
(Wa)DIQaM-NR for four selected distortion types from TID2013 (Gaussian blur, JPEG compression, JP2K compression and local block-wise distortions of different
intensity). While for (WA)DIQaM-FR we see the same behavior on single distortion types
as on aggregated distortion types, in the \ac{NR} \ac{IQA} case weighted patch aggregation
pooling decreases performance for GB, JPEG and JP2K, but increases performance for LBDDI. 
We conjecture that for most distortions information from the reference image is crucial to assign
local weights for pooling, but if distortions are strongly inhomogeneous as LBDDI, the
distorted image is sufficient to steer weighting.
One of the reasons for CLIVE being so challenging
for \ac{IQA} is that distortions and scenes are spatially much more inhomogeneous than in LIVE or
TID2013, which the weighted patch aggregation can compensate for.
This also explains the huge increase in performance by weighted patch
aggregation pooling for the \textit{exotic} subset of the TID2013 in
\Tabref{tab:performance_by_dist_type_tid2013} as this subset contains a larger amount
of inhomogeneous distortion types.
The resulting local weights will be examined more closely in the next section.

\subsection{Local Weights}
\label{ssec:local_weights}
\begin{figure}[b!]
  \centering
  \subfloat[Distorted Image]{
  \includegraphics[width=0.23\textwidth]{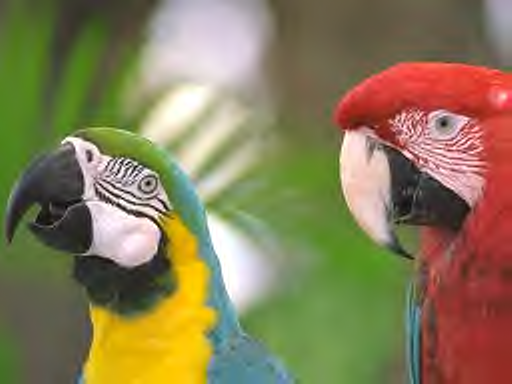}
    
  }
  
  \subfloat[$y_i$, DIQaM-FR ]{
  \includegraphics[width=0.23\textwidth]{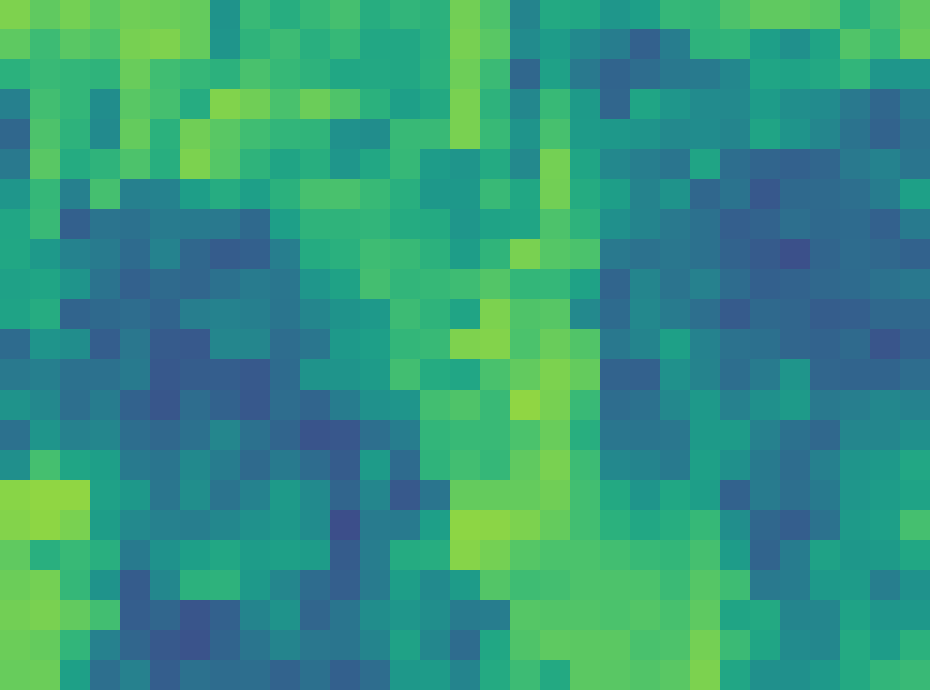}
  \label{fig:jp2k_y_fr_pw}
  }
  \subfloat[$y_i$, DIQaM-NR]{
  \includegraphics[width=0.23\textwidth]{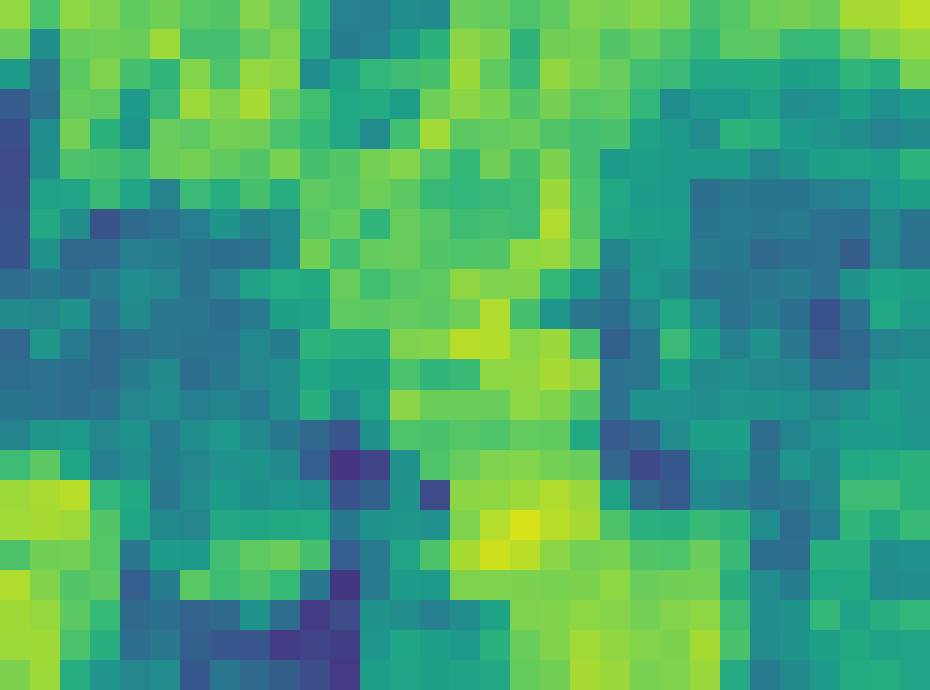}
  \label{fig:jp2k_y_nr_pw}
  }
  
  \subfloat[$y_i$, WaDIQaM-FR]{
  \includegraphics[width=0.23\textwidth]{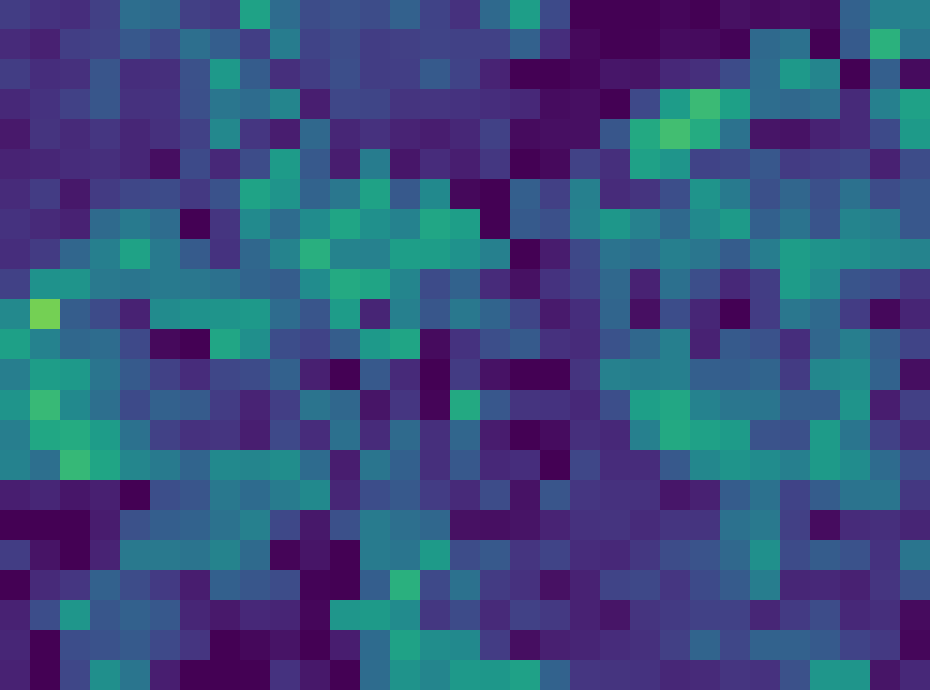}
  \label{fig:jp2k_y_fr_weighted}
  }
  \subfloat[$y_i$, WaDIQaM-NR]{
  \includegraphics[width=0.23\textwidth]{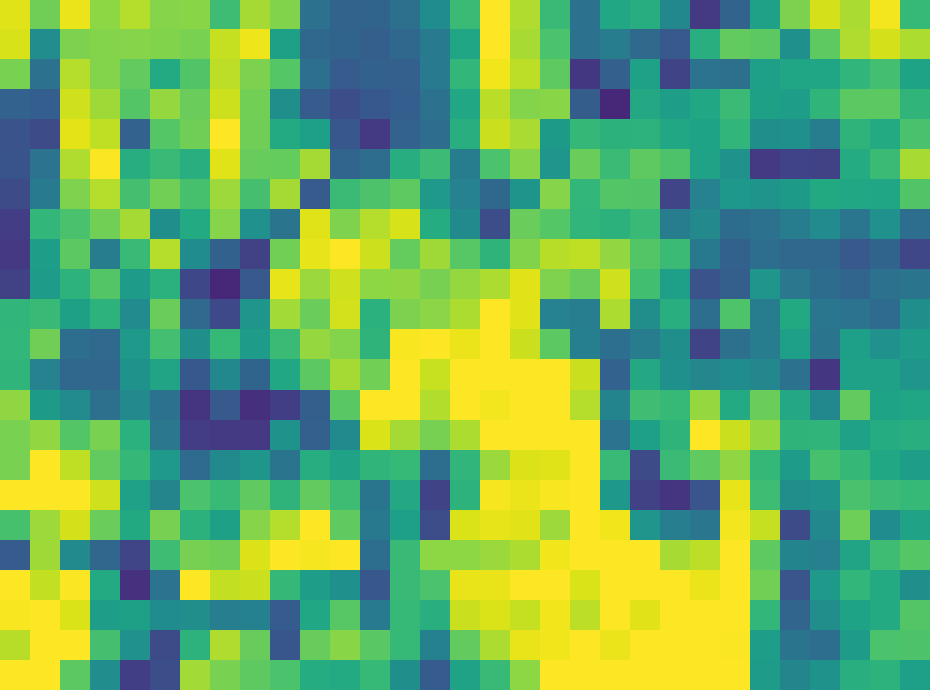}
  \label{fig:jp2k_y_nr_weighted}
  }
  
  \subfloat[$\alpha^*_i$, WaDIQaM-FR]{
  \includegraphics[width=0.23\textwidth]{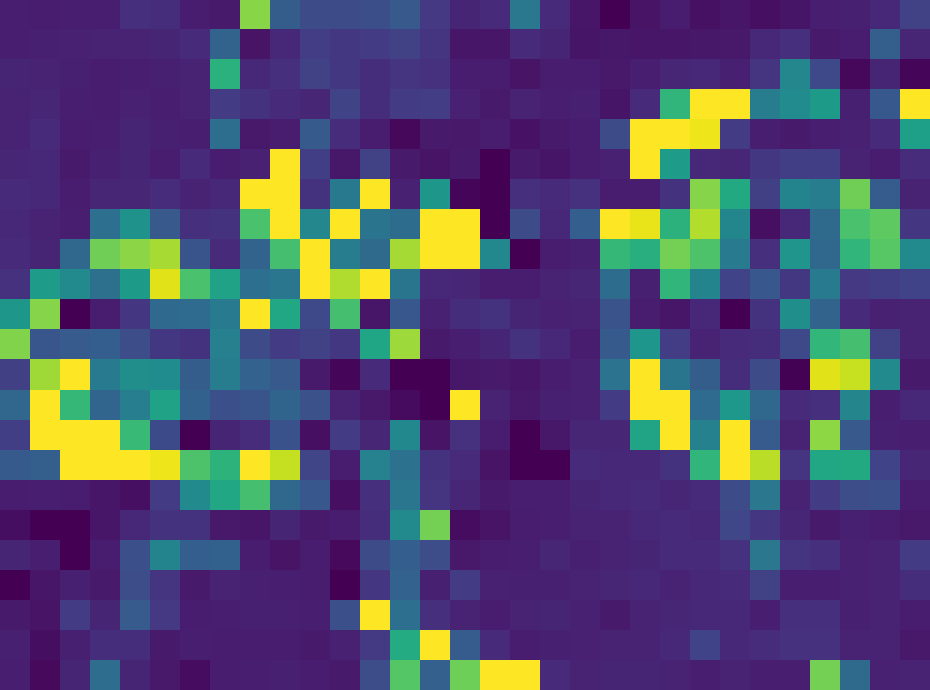}
  \label{fig:jp2k_a_fr_weighted}
  }
  \subfloat[$\alpha^*_i$, WaDIQaM-NR]{
  \includegraphics[width=0.23\textwidth]{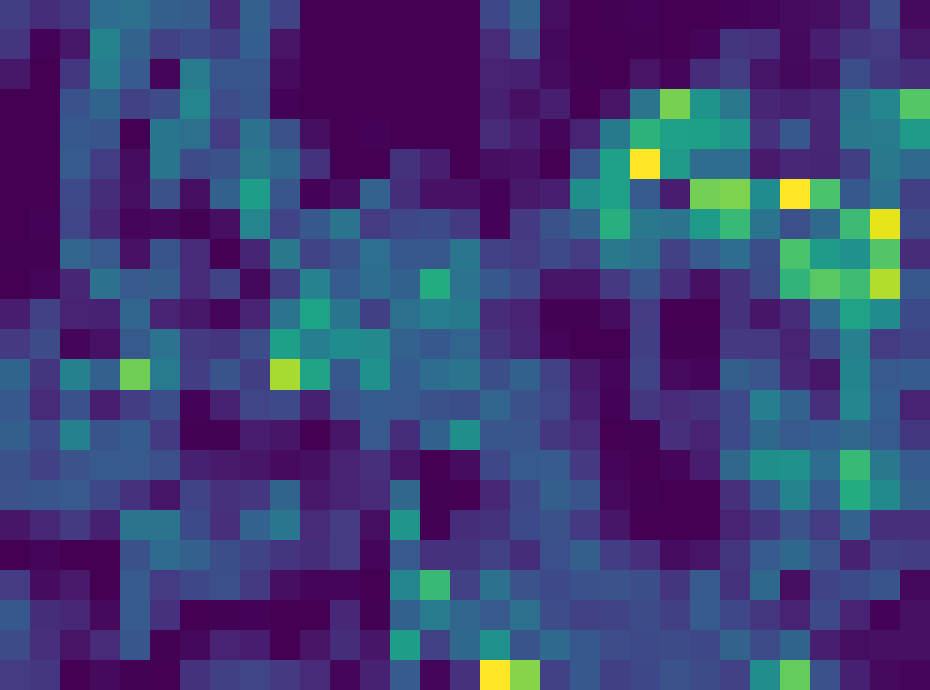}
  \label{fig:jp2k_a_nr_weighted}
  }
\caption{Local qualities $y_i$ and weights $\alpha^*_i$ for a JP2K distorted image from TID2013.
Blue indicates low, yellow high values of local distortions and weights, respectively. The MOS value is 34, predicted qualities
are 54 by DIQaM-FR, 42 by WaDIQAM-FR, 60 by DIQaM-NR, and 70 by WaDIQam-NR.}
\label{fig:local_maps_jp2k}
\end{figure}
\begin{figure}[b!]
  \centering
  \subfloat[Distorted Image]{
  \includegraphics[width=0.23\textwidth]{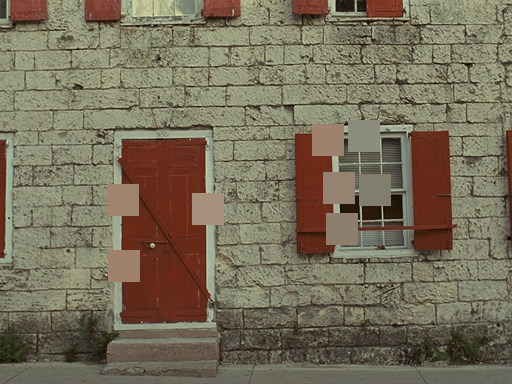}
  
  }
  
  \subfloat[$y_i$, DIQaM-FR ]{
  \includegraphics[width=0.23\textwidth]{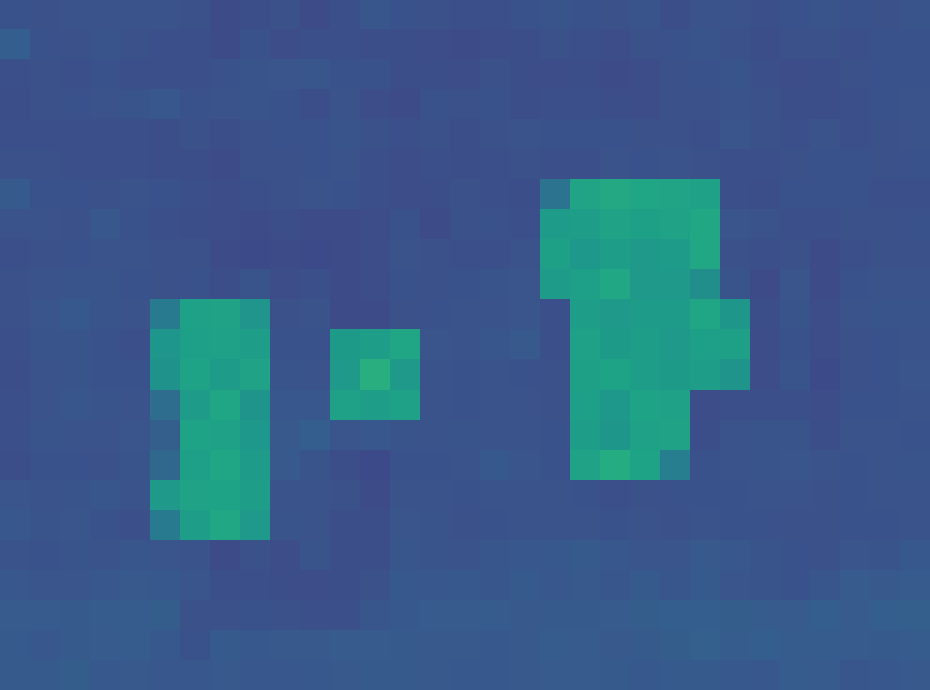}
  \label{fig:lbddi_y_fr_pw}
  }
  \subfloat[$y_i$, DIQaM-NR]{
  \includegraphics[width=0.23\textwidth]{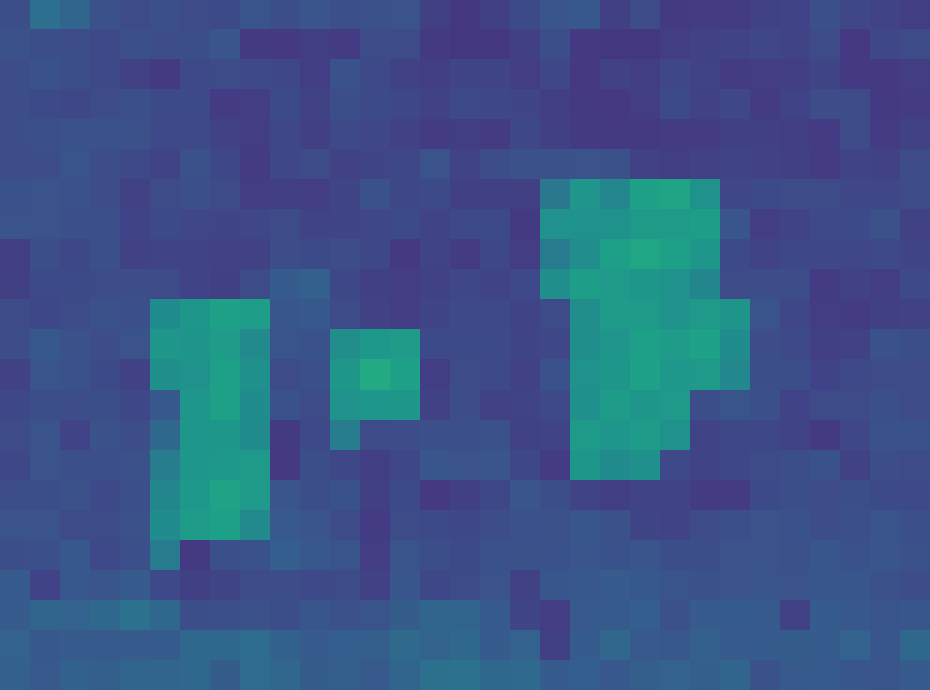}
  \label{fig:lbddi_y_nr_pw}
  }
  
  \subfloat[$y_i$, WaDIQaM-FR]{
  \includegraphics[width=0.23\textwidth]{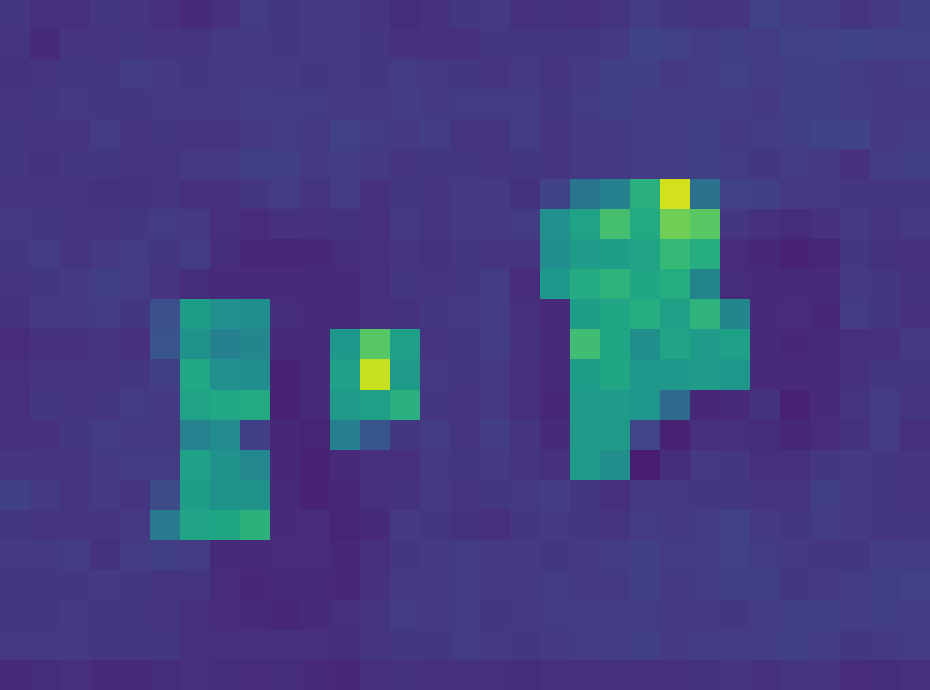}
  \label{fig:lbddi_y_fr_weighted}
  }
  \subfloat[$y_i$, WaDIQaM-NR]{
  \includegraphics[width=0.23\textwidth]{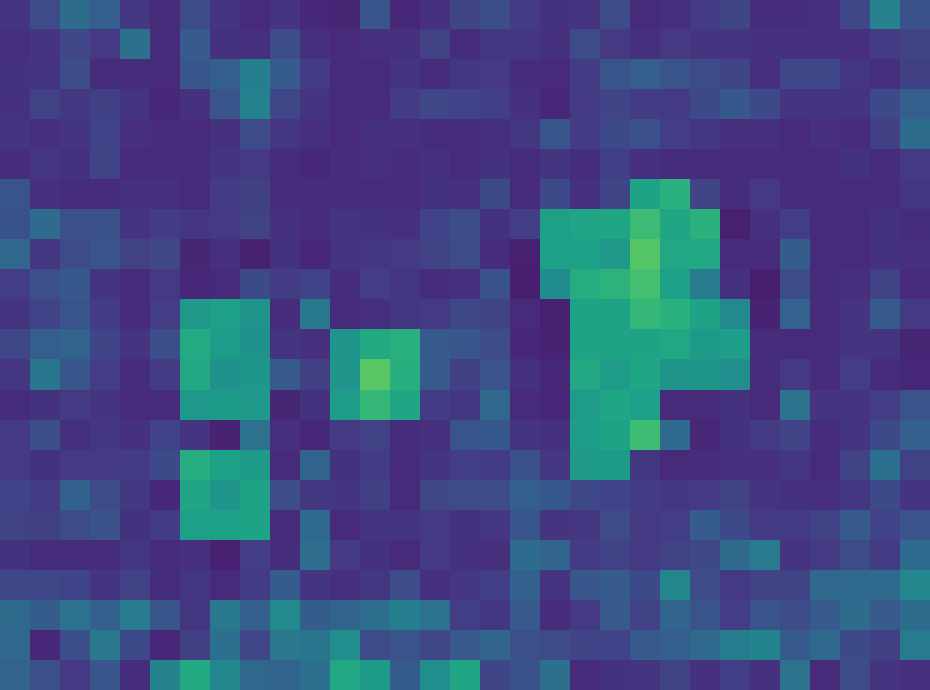}
  \label{fig:lbddi_y_nr_weighted}
  }
  
  \subfloat[$\alpha^*_i$, WaDIQaM-FR]{
  \includegraphics[width=0.23\textwidth]{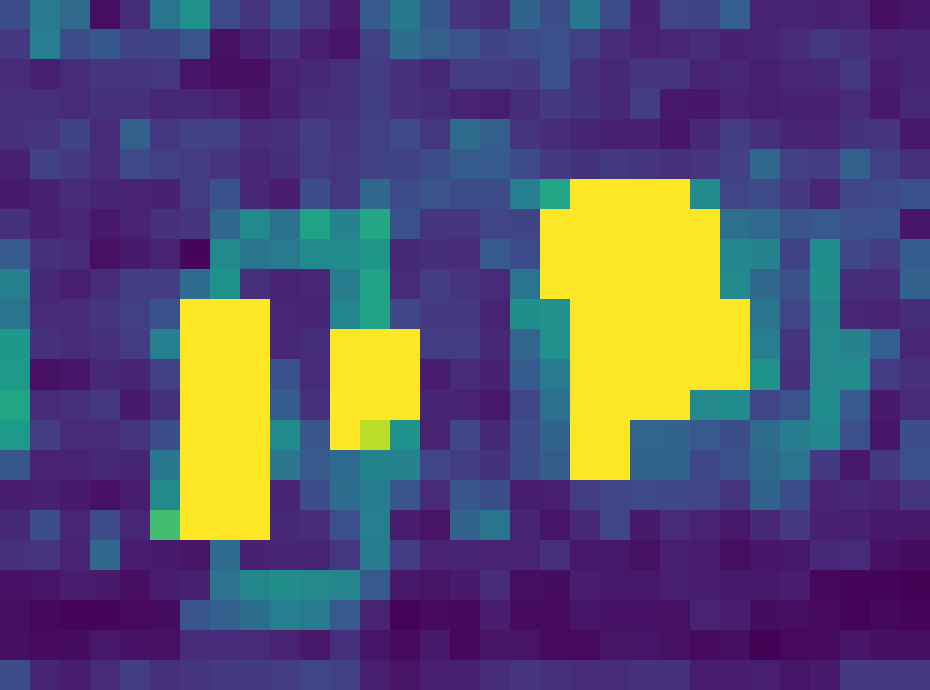}
  \label{fig:lbddi_a_fr_weighted}
  }
  \subfloat[$\alpha^*_i$, WaDIQaM-NR]{
  \includegraphics[width=0.23\textwidth]{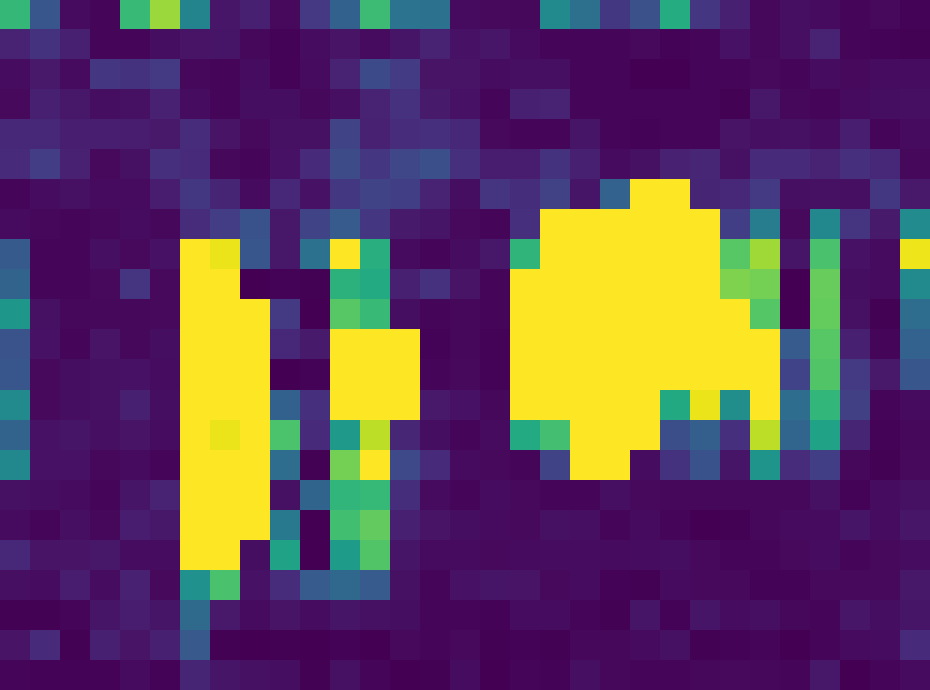}
  \label{fig:lbddi_a_nr_weighted}
  }
\caption{Local qualities $y_i$ and weights $\alpha^*_i$ for a LBDDI distorted image from TID2013.
Blue indicates low, yellow high values of local distortions and weights, respectively. The MOS value is 59, predicted qualities are 30 by DIQaM-FR,
51 by WaDIQAM-FR, 27 by DIQaM-NR, and 53 by WaDIQam-NR.
}
\label{fig:local_maps_lbdi}
\end{figure}

\begin{figure}[htb!]
  \centering
  \subfloat[Input Image]{
  \includegraphics[width=0.23\textwidth]{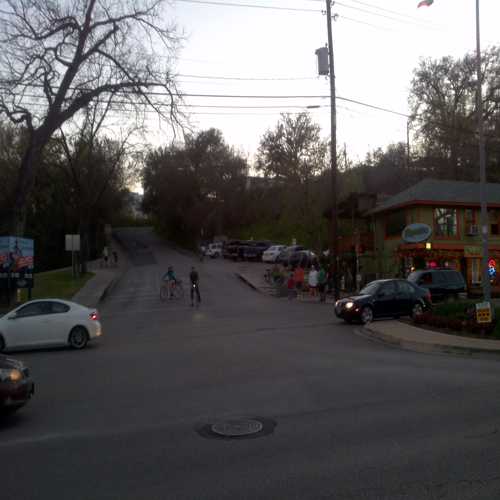}
  \label{fig:cdb_fail_orig}
  }
  \subfloat[Input Image]{
  \includegraphics[width=0.23\textwidth]{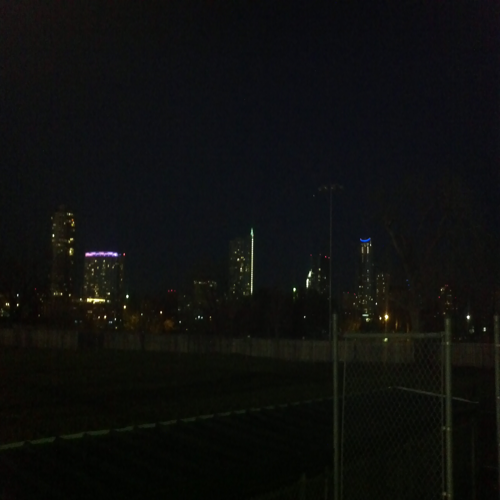}
  \label{fig:cdb_succ_orig}
  }
    
  \subfloat[$y_i$, DIQaM-NR]{
  \includegraphics[width=0.23\textwidth]{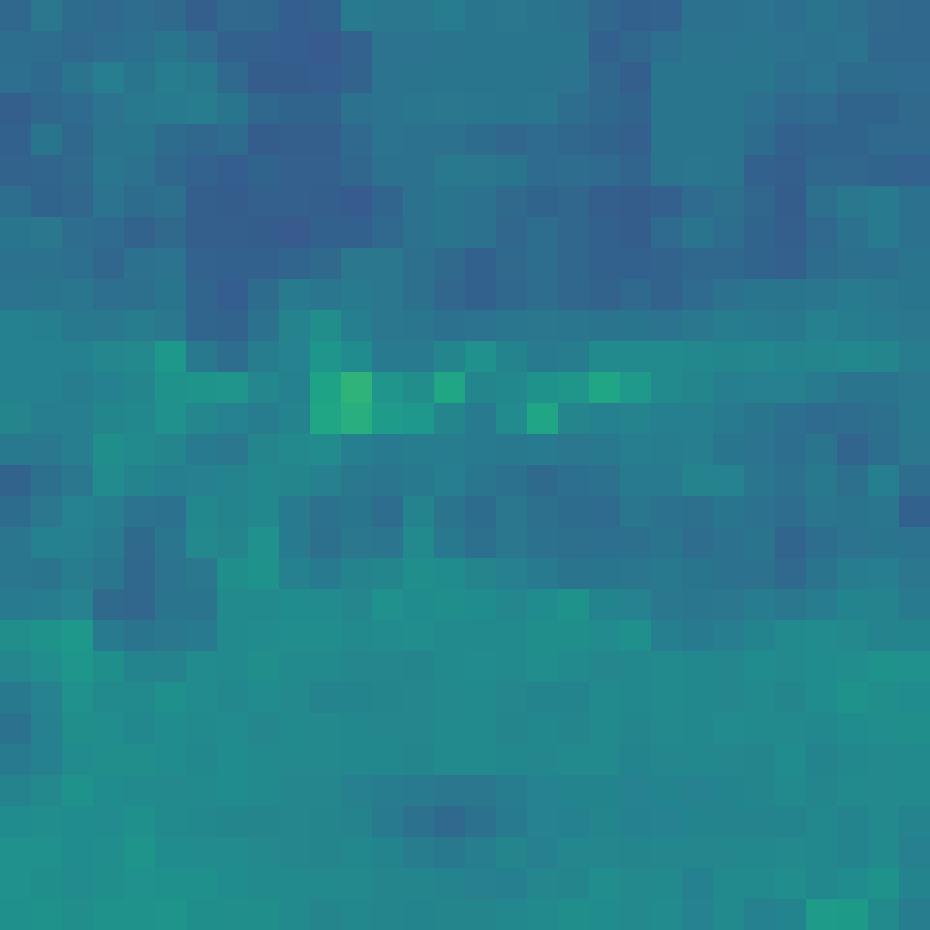}
  \label{fig:cdb_fail_pw_y}
  }
  \subfloat[$y_i$, DIQaM-NR]{
  \includegraphics[width=0.23\textwidth]{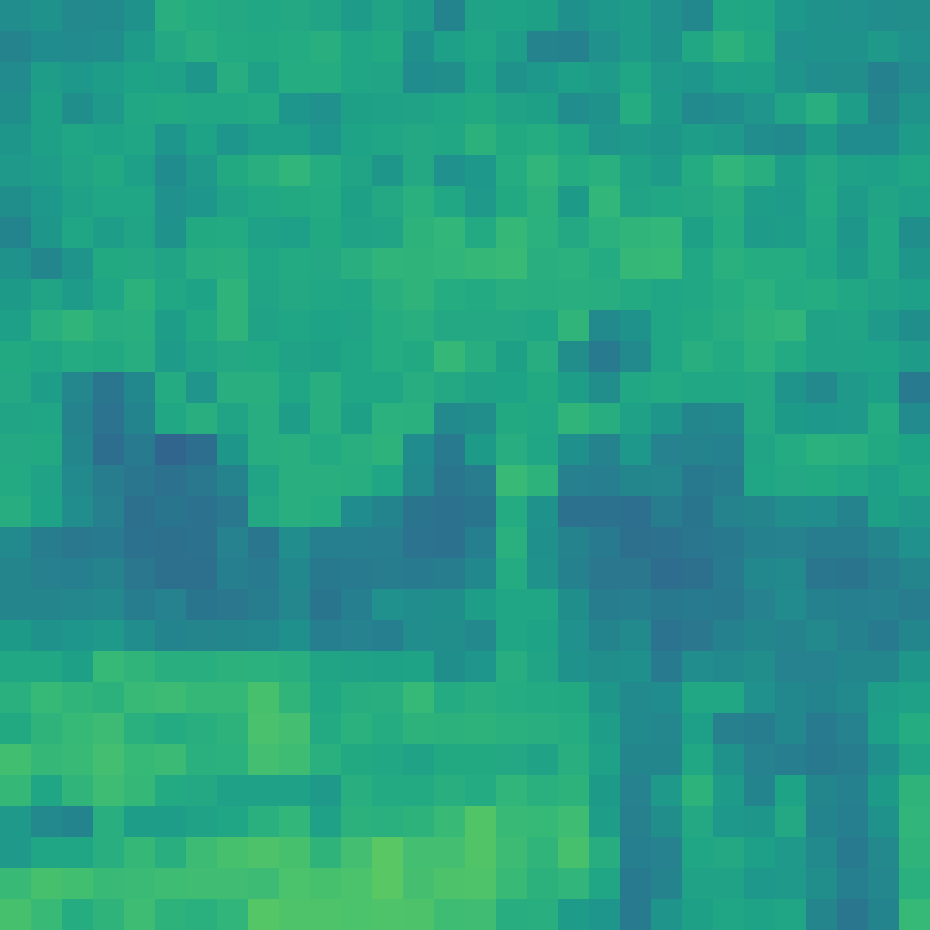}
  \label{fig:cdb_succ_pw_y}
  }
  
  \subfloat[$y_i$, WaDIQaM-NR]{
  \includegraphics[width=0.23\textwidth]{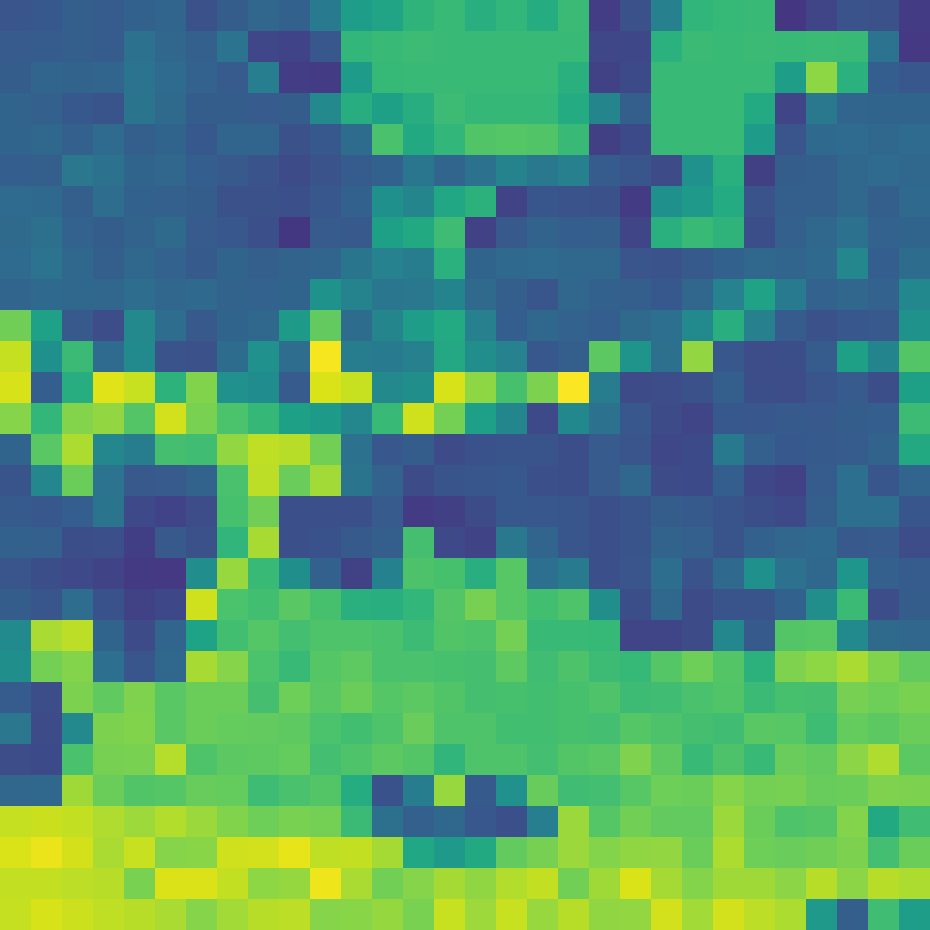}
  \label{fig:cdb_fail_weight_y}
  }
  \subfloat[$y_i$, WaDIQaM-NR]{
  \includegraphics[width=0.23\textwidth]{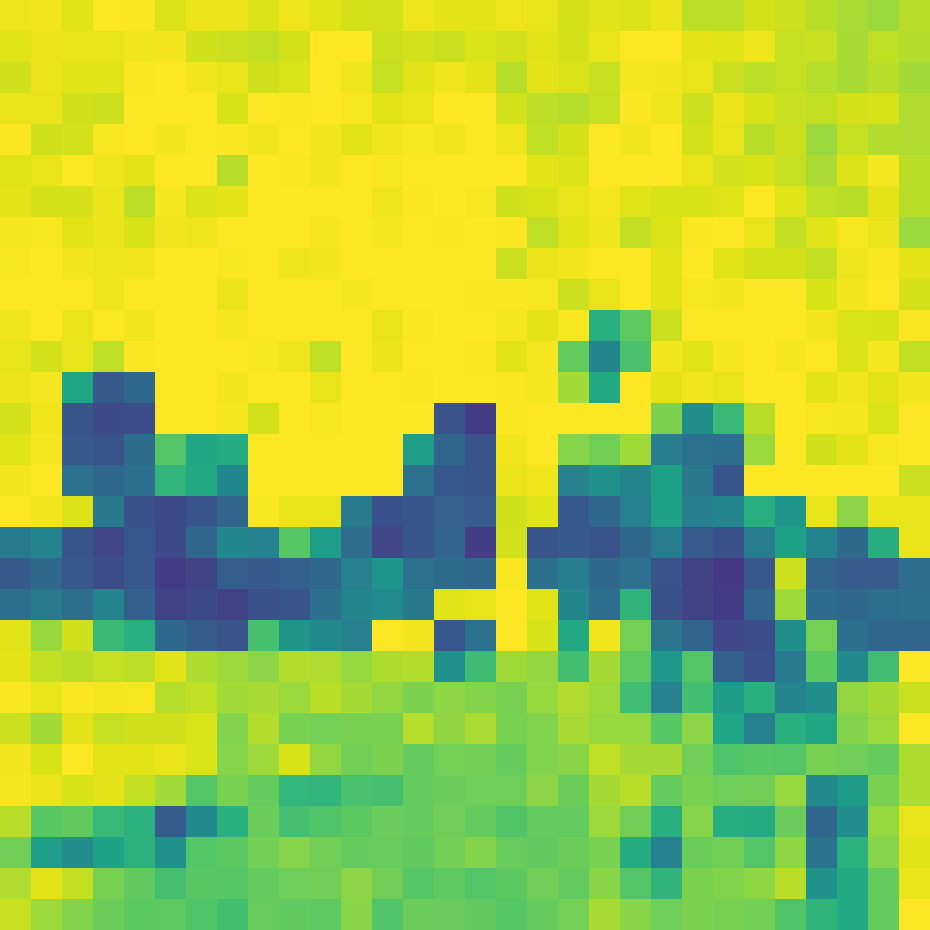}
  \label{fig:cdb_succ_weight_y}
  }
  
  \subfloat[$\alpha^*_i$, WaDIQaM-NR]{
  \includegraphics[width=0.23\textwidth]{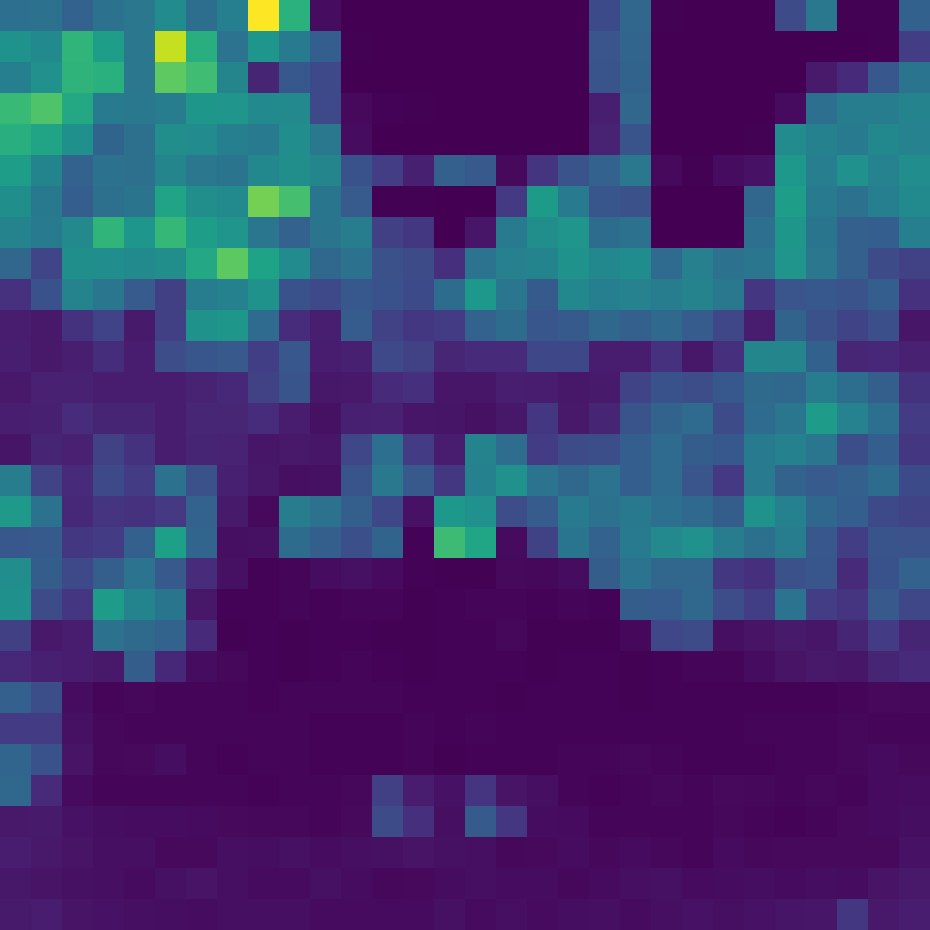}
  \label{fig:cdb_fail_weight_a}
  }
  \subfloat[$\alpha^*_i$, WaDIQaM-NR]{
  \includegraphics[width=0.23\textwidth]{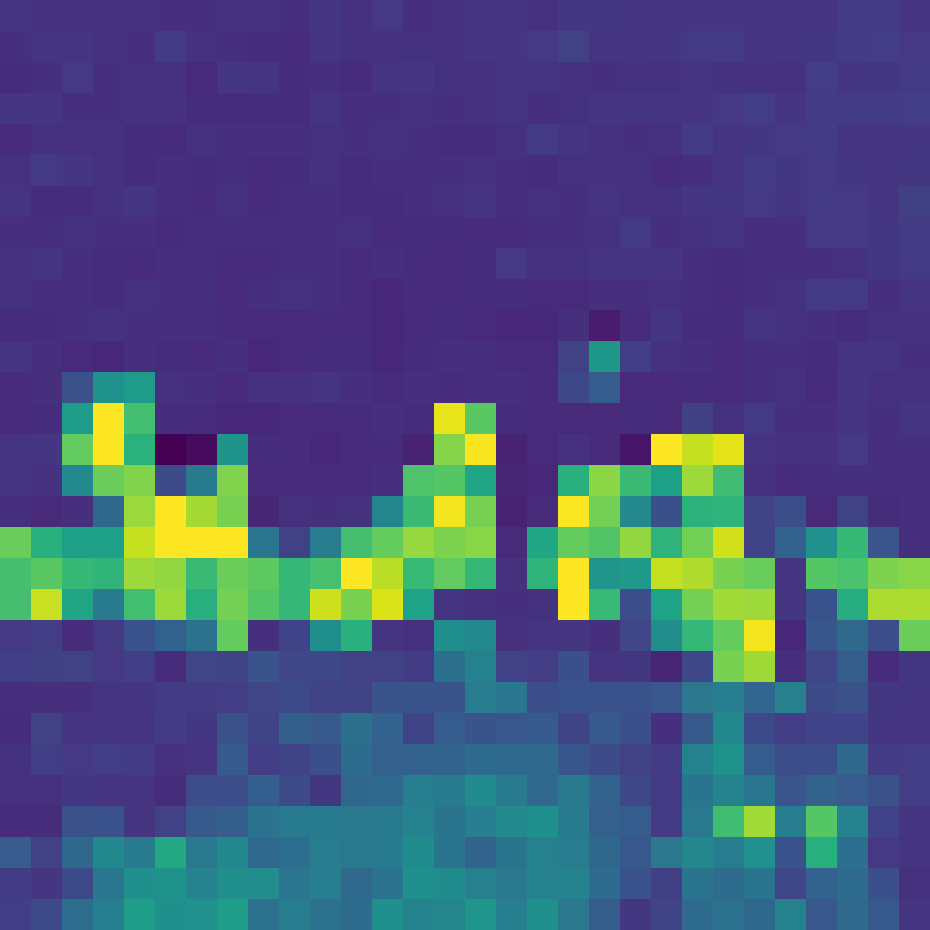}
  \label{fig:cdb_succ_weight_a}
  }
\caption{Local qualities $y_i$ and weight maps $\alpha^*_i$ for two image from
CLIVE. 
Left column: MOS value is 43, predicted qualities are 42 by DIQaM-NR, 34 by
WaDIQam-NR.
Right column: MOS value is 73, predicted qualities are 56 by DIQaM-NR, 66 by
WaDIQam-NR.}
\label{fig:local_maps_clive}
\end{figure}

%
%
%

The previous sections showed that the weighted average patch aggregation scheme has an influence
that depends on the distortion type and the availability of a
reference.

\Figref{fig:local_maps_jp2k} shows the local qualities $y_i$ and weights
$\alpha^*_i$  for an image subject
to JP2K compression from TID2013.
The MOS value of the distorted image is 34; the relation between prediction accuracies of the four
different models are as expected from the previous evaluations (DIQaM-FR: 54,
WaDIQaM-FR: 42, DIQaM-NR:60, WaDIQaM-NR: 70). The left column shows the quality
and weight maps computed by the proposed \ac{FR} models, the right column the maps from the \ac{NR}
models. The DIQaM-FR/NR assign higher distortion values to the background of
the image than to the two foreground objects (Figs.~\ref{fig:jp2k_y_fr_pw}
and \ref{fig:jp2k_y_nr_pw}).
In the \ac{FR} case, the local weights provide some rough image segmentation
as higher weights are assigned to image regions containing objects
(\Figref{fig:jp2k_a_fr_weighted}). This fails in the \ac{NR} case (\Figref{fig:jp2k_a_nr_weighted}),
which explains the performance drop from  DIQaM-NR to WaDIQaM-NR observed in
\Subsubsecref{sssec:nr_iqa_single_database_performance}.

The local quality and weight maps resulting from an image subject to
spatially highly variant distortions, in this example LBDDI from TID2013, is shown in
\Figref{fig:local_maps_lbdi}. Here, for WaDIQaM-FR as well as for WaDIQaM-NR
the network is able to assign higher weights to the distorted image regions and
by that improve prediction accuracy compared to the models employing simple
average pooling.
Note that, as in \Figref{fig:local_maps_jp2k}, WaDIQaM-FR is
again able to roughly segment the image, whereas WaDIQaM-NR again fails
at segmentation.
However, for this extreme distortion type the structure of the image is of less 
importance. 

In \Subsubsecref{sssec:nr_iqa_single_database_performance} we conjectured that one
reason for WaDIQaM-NR to improve prediction performance over DIQaM-NR for
CLIVE, but to decrease performance on LIVE and TID2013 is the higher amount of spatial
variance in CLIVE. 
\Figref{fig:local_maps_clive} exemplifies this effect for two images from CLIVE. 

The top row shows the test images, where the left-hand sided one 
(\Figref{fig:cdb_fail_orig}) suffers rather globally from underexposure,
rendering identification of a certain area of higher impairment difficult, while
the right-hand sided one (\Figref{fig:cdb_succ_orig}) contains clear regions of
interest that are rather easy to identify against the black background.
The lower rows show the corresponding quality and weight maps.
\Figref{fig:cdb_succ_weight_a} shows that for this spatially highly concentrated
scene, WaDIQaM-NR is able to identify the patches contributing the most to the overall image structure. However, as \Figref{fig:cdb_fail_weight_a} shows, it fails to do so for the homogeneously impaired image.

Another important observation from 
\Figref{fig:local_maps_jp2k}, \Figref{fig:local_maps_lbdi} and \Figref{fig:local_maps_clive} is
that weighted average patch aggregation has an influence also on the quality maps. Thus, the
joint  optimization introduces an interaction between $y_i$ and $\alpha^*_i$
that is adaptive to the specific distortion. 

\subsection{Cross-Database Evaluation}
\label{ssec:cross_dataset_eval}
\subsubsection{Full-Reference Image Quality Assessment}
\label{sssec:cross_dataset_fr}

\begin{table}[htb!]
\centering
\caption{SROCC Comparison in Cross-Database Evaluation.
All Models are Trained on Full LIVE or TID2013, respectively, and
Tested on either CSIQ, LIVE or TID2013.}
\label{tab:comparison_fr_crossdata}
    \begin{tabular}{  l | c | c | c | c }
    
  Trained on:           & \multicolumn{2}{c|}{LIVE}
  &\multicolumn{2}{c}{TID2013}
  \\\hline
  Tested on: & TID2013 & CSIQ & CSIQ & LIVE \\
   \hline
  DOG-SSIM \cite{Pei2015}      & \bf{0.751}  & \bf{0.914}  & 0.925 & \bf{0.948}\\
  DIQaM-FR (proposed) & 0.437 & 0.660 & 0.863 & 0.796\\
  WaDIQaM-FR (proposed)  & \bf{0.751}  & 0.909  & \bf{0.931} & 0.936\\
    
    \end{tabular}
\end{table}

\Tabref{tab:comparison_fr_crossdata}
shows the  results for models trained on LIVE and tested on TID2013 and CSIQ,
and for models  trained on TID2013 and tested on LIVE and CSIQ. Results are compared
to  DOG-SSIM, as most other \ac{FR} \ac{IQA} methods compared to do not rely on training.
In all combinations of training and test set the DIQaM-FR model shows insufficient generalization
capabilities, while WaDIQaM-FR performs  best among the two proposed spatial
pooling schemes and comparable to DOG-SSIM.
The superior results of the model trained on TID2013 over the model trained on LIVE when tested
on CISQ indicate that a larger training set may lead to better generalization.

\subsubsection{No-Reference Image Quality Assessment}
\label{sssec:cross_dataset_nr}

\begin{table}[htb!]
\begin{center}
\caption{SROCC in Cross-Database Evaluation from. All models
are Trained on the Full LIVE Database and Evaluated on CSIQ and TID2013. The
Subsets of CSIQ and TID2013 Contain only the 4 distortions Shared with
LIVE.}
    \begin{tabular}{  l | c | c | c | c |}
    
   & \multicolumn{2}{c|}{subset}& \multicolumn{2}{c|}{full}\\
   \cline{2-5}
   Method & CSIQ  & TID2013 & CSIQ  &  TID2013  \\
      \hline
  DIIVINE \cite{Moorthy2011}      & - & - & 0.596   &
  0.355
  \\
  BLIINDS-II \cite{Saad2012}      & -    & -   & 0.577&
  0.393 \\
  BRISQUE \cite{Mittal2012}       & 0.899  & 0.882   &
  0.557 & 0.367 \\
  CORNIA \cite{Ye2012a}           & 0.899 & 0.892 & 0.663
  & 0.429 \\
  QAF \cite{zhang2014training}        & - & - & 0.701  &
  0.440 \\
  CNN \cite{Kang2014}    & - & 0.920 & -  & - \\
  SOM \cite{Zhang2015}   & - & \bf{0.923} & -  & - \\
  DIQaM-NR  (proposed) & \bf{0.908} & 0.867 & 0.681  &
  0.392
  \\
  WaDIQaM-FR  (proposed) & 0.866  & 0.872&  \bf{0.704}
  &\bf{0.462} \\
    
    \end{tabular}
\label{tab:comparison_nr_crossdata}
\end{center}
\end{table}

\begin{table}[htb!]
\begin{center}
\caption{SROCC Comparison in Cross-Database Evaluation.
All Models are Trained on the full TID2013 Database and Evaluated on CSIQ.}
    \begin{tabular}{  l | c }
    
  Method           & CSIQ full  \\
    \hline
  DIIVINE \cite{Moorthy2011}      & 0.146\\
  BLIINDS-II \cite{Saad2012}      & 0.456 \\
  BRISQUE \cite{Mittal2012}       & 0.639 \\
  CORNIA \cite{Ye2012a}           & 0.656 \\
  DIQaM-NR (proposed) & 0.717 \\
  WaDIQaM-FR  (proposed)  & \bf{0.733} \\

    \end{tabular}
\label{tab:comparison_nr_tid_crossdata}
\end{center}
\end{table}

For evaluating the generalization ability of the proposed \ac{NR} \ac{IQA} models, we extend
cross-database experiments presented in \cite{zhang2014training} with our results.
For that, a model trained on the full LIVE database is evaluated on subsets of CSIQ and TID2013,
containing only the four distortions types shared between the three databases (JPEG, JP2K,
Gaussian blur and white noise).
Results are shown in \Tabref{tab:comparison_nr_crossdata}. 
While DIQaM-NR shows superior performance compared to BRISQUE and  CORNIA  on the
CISQ subset, the proposed models are outperformed by the other
state-of-the-art  methods when cross-evaluated on the subset of TID2013. 
As for the full CSIQ database, the two  unseen distortions (i.e. pink additive noise and contrast
change) are considerably  different in their visual appearance compared to the ones considered during training.
Thus, it  is not surprising that all compared IQA methods perform  worse in
this setting.
Despite performing worse on the single database experiments, WaDIQaM-NR seems to be able to adapt
better to unseen distortions than DIQaM-NR. This is in line with the results on CLIVE, as for CLIVE
a specific picture's mixture of distortions is less likely to be in the training set than e.g. for
LIVE.
Although being a vague comparison as TID2008 contains less distorted images per distortion type,
is it worth to note that BIECON obtains a SROCC=0.923 in a similar experiment (trained on LIVE,
tested on the 4 distortions types of the smaller TID2008 and excluding one image).

Given the relatively wide variety of distortions types in TID2013 and with only 4 out of 24
distortions being contained in the training set, a model trained on LIVE  can be expected
to perform worse if tested on TID2013 than if tested on CSIQ.
Unsurprisingly, none of the learning-based
methods available for comparison is able to achieve a SROCC over 0.5. These
results suggest that learning a truly  non-distortion-specific IQA metric using
only the examples in the LIVE database is hard  or even impossible.
Nevertheless, the proposed methods obtain competitive, but still very unsatisfactory results.

\Tabref{tab:comparison_nr_tid_crossdata} shows the performance in terms of SROCC for models trained
on full TID2013 and tested on full CSIQ. Performance of 
DIIVINE, BLIINDS-II and CORNIA trained on TID2013 is decreased compared to being trained on
LIVE, despite TID2013 being the larger and more diverse training set, while BRISQUE and the proposed
models are able to make use of the larger training set size.
This follows the notion that the generalization capabilities
of  deep neural  networks depend on the size and diversity of the training set.

Even though the proposed  methods outperform comparable methods, a SROCC of
0.733 on the CSIQ  dataset is still far from being satisfactory. Despite having
more  images in total and more distortions  than LIVE, the TID2013 has 4
reference images fewer. Thus, training on TID2013  has the same  short-comings
as training on LIVE when it comes to adaption to unseen images.

Note that for the evaluation of learning based IQA methods databases are split into training, testing and (for some approaoches) validation sets.
Models are the trained and evaluated for a number of different random splits and the performances of these splits need to be pooled for comparison.
This renders evaluation a random process. Performances of different models as reported in the literature are obtained based on different split ratios,
different numbers of splits and different ways of pooling. This may have a slight influence on the comparison competing methods.

\subsection{Convergence Evaluation}
Results presented in the previous sections were obtained when $N_p=32$ patches are considered for
quality estimation. However, the prediction performance and accuracy depends on the number of patches used.
This subsection examines the influence of $N_p$ and justifies the choice of $N_p=32$.

\label{ssec:convergence_evaluation}
\subsubsection{Full-Reference Image Quality Assessment}
\begin{figure}[htb!]
\centering
\subfloat[LIVE]{
\includegraphics[width=0.23\textwidth]{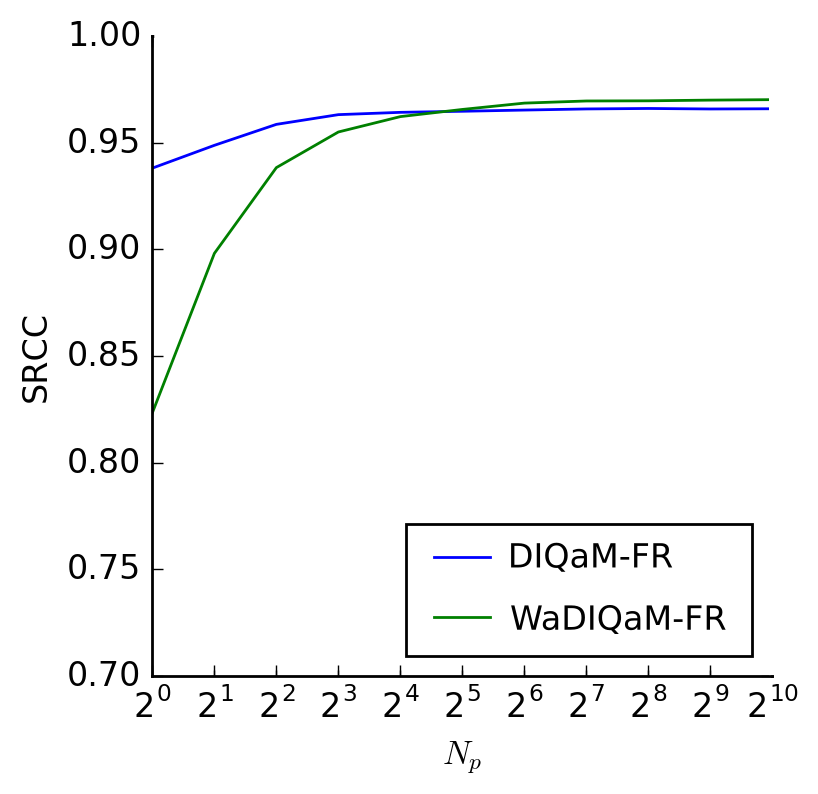}
}
\subfloat[TID2013]{
\includegraphics[width=0.23\textwidth]{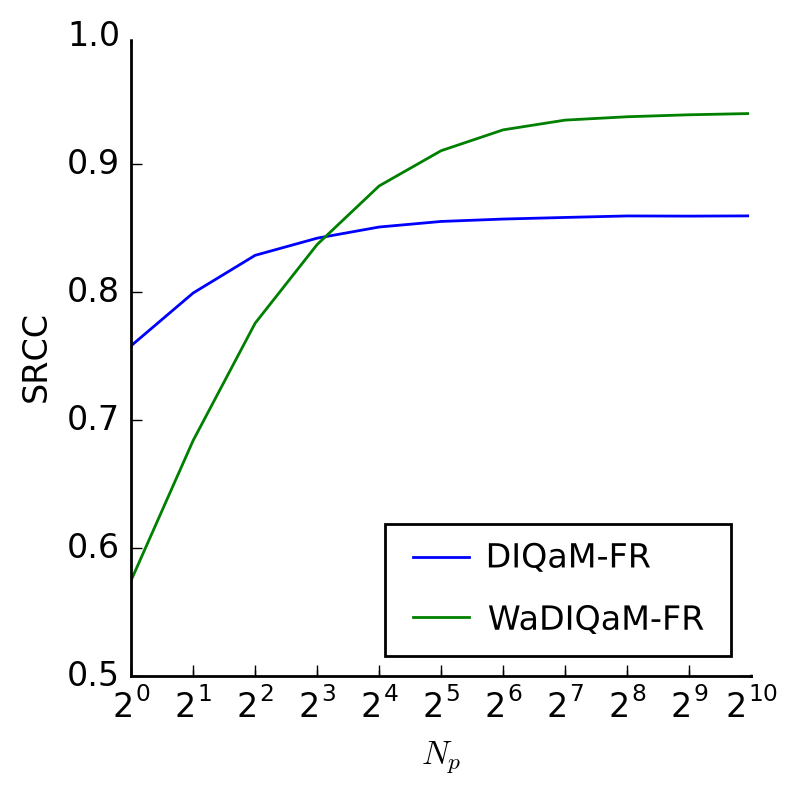}
}

\caption{SROCC of the proposed CNN for FR IQA in dependence of the number of randomly sampled patches evaluated on LIVE and TID2013.}
\label{fig:fr_numberOfPatchesVsPerformance}
\end{figure}
\Figref{fig:fr_numberOfPatchesVsPerformance} plots the performance for the
models trained and tested on LIVE and TID2013 in dependence of $N_p$ in terms of SROCC and shows a
monotonic increase of performance towards saturation for  larger $N_p$.
As noted before, weighted average patch aggregation improves the prediction performance over simple
averaging, here we see that this holds only for a sufficient number of patches $N_p>8$.
The WaDIQaM-FR saturates at the maximum performance with $N_p\approx 32$,
whereas the DIQaM-FR saturates already at $N_p\approx 16$ in all three evaluation metrics.

\subsubsection{No-Reference Image Quality Assessment}
\begin{figure}[htb!]
  \centering
\subfloat[LIVE]{
\includegraphics[width=0.23\textwidth]{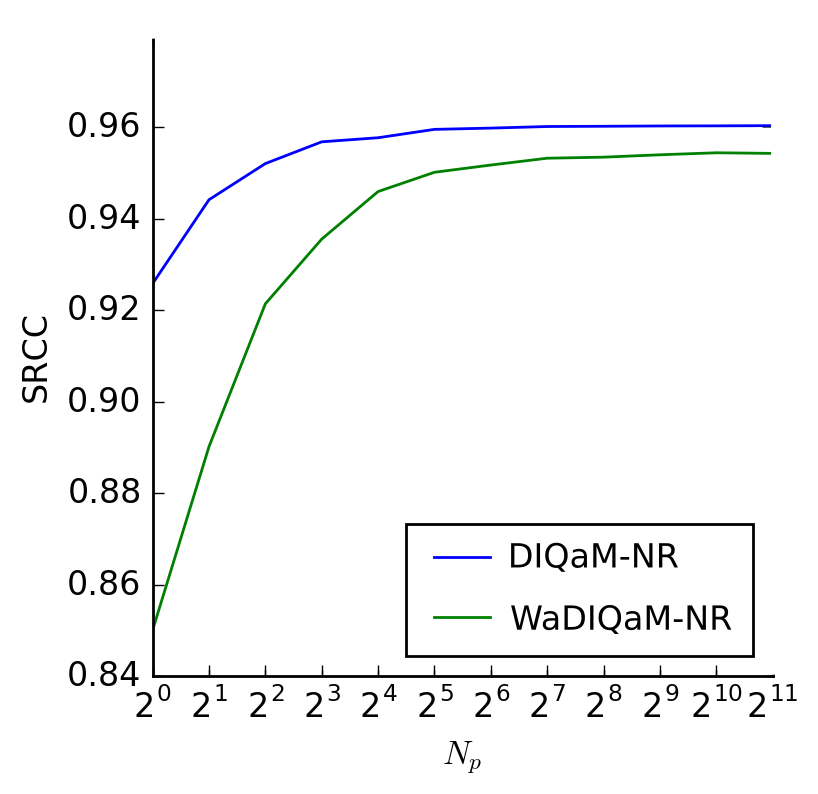}
}
\subfloat[TID2013]{
\includegraphics[width=0.23\textwidth]{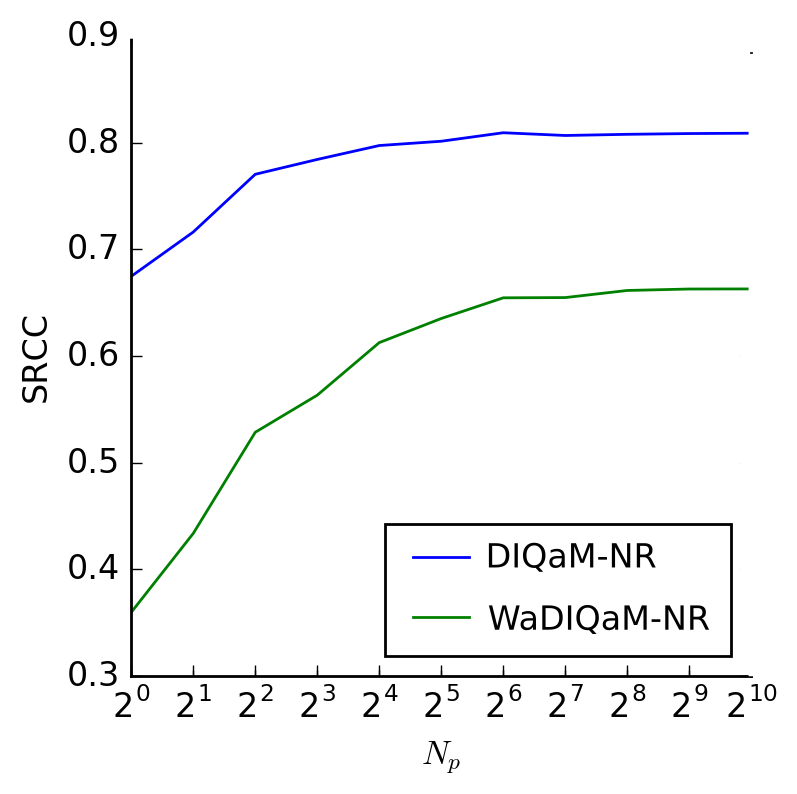}
}
 \caption{SROCC of the proposed CNN for NR IQA in dependence of the number of randomly sampled
 patches evaluated on LIVE and TID2013.}
  \label{fig:nr_numberOfPatchesVsPerformance}
\end{figure}

\begin{figure}[t]
\centering
\subfloat[DIQaM-NR]{
\includegraphics[width=0.23\textwidth]{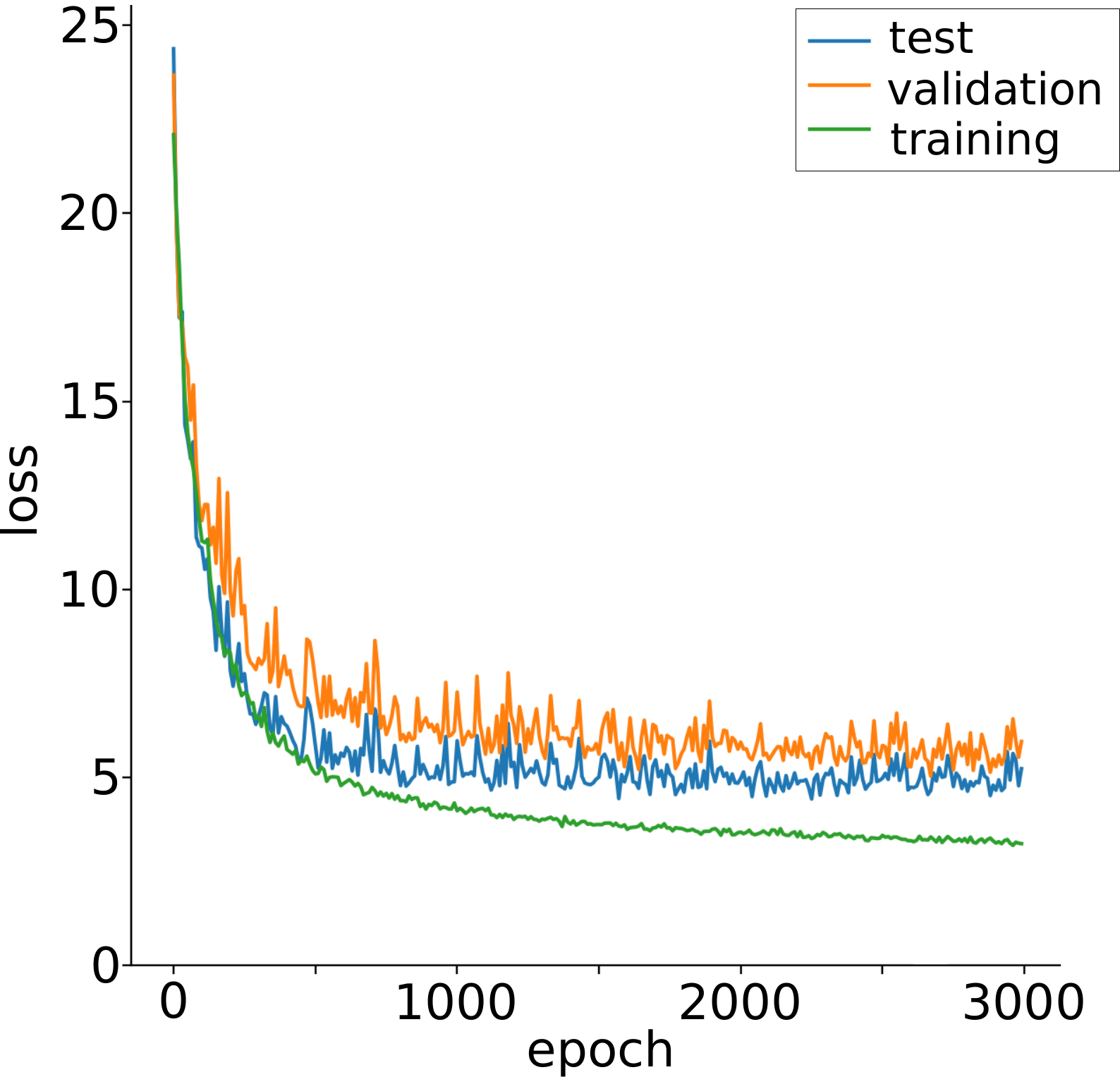}
}
\subfloat[WaDIQaM-NR]{
\includegraphics[width=0.23\textwidth]{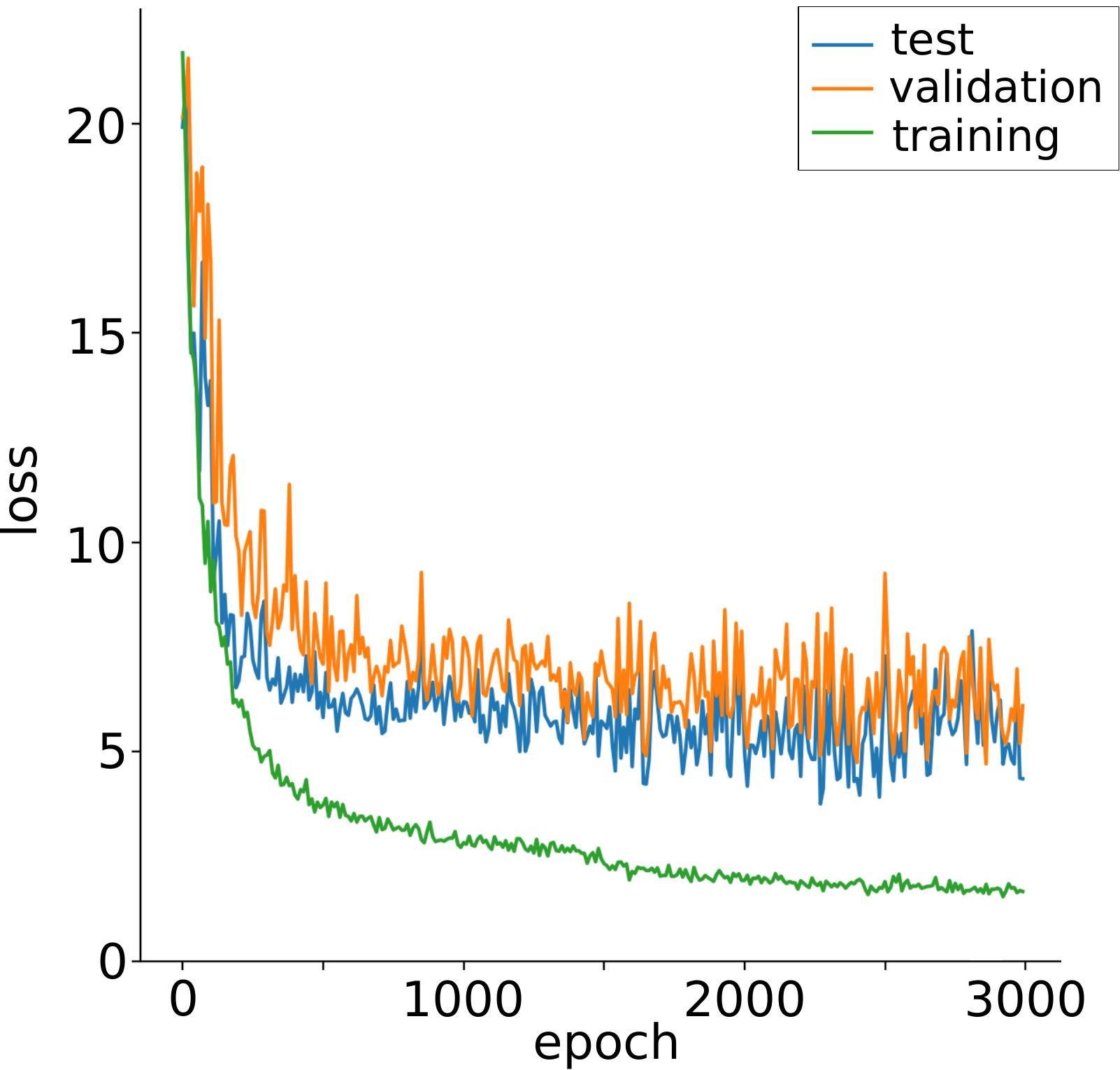}
}
\caption{Loss in training, validation and testing vs. number of epochs in training for DIQaM-NR
and WaDIQaM-NR.}
\label{fig:loss_vs_epochs}
\end{figure}
The influence  of the number of  patches $N_p$ on 
the prediction performance of DIQaM-NR and WaDIQaM-NR is plotted in 
\Figref{fig:nr_numberOfPatchesVsPerformance}. 
For both pooling methods and on both databases SROCC improves
monotonically with increasing number of patches $N_p$ until saturation.

On LIVE, for DIQaM-NR  saturation sets in at about $N_p\approx 16$ to reach
its maximal performance, whereas WaDIQaM-NR  reaches its maximal performance at $N_p\approx
256$. 
Over the whole range of $N_p$ the performance of average patch aggregation
is superior to the performance of weighted average patch aggregation and the
difference is largest for small numbers $N_p$. 
This is because the  weighted average acts as a filter that ignores patches with lower weights and
thereby reduces the number of patches considered in quality estimation.
Qualitatively the
same results are obtained on TID2013.

\Figref{fig:loss_vs_epochs} shows the course of optimization loss (effectively the \ac{MAE})
during training, validation and testing in dependence of the number of epochs of training
exemplified for DIQaM-NR and WaDIQaM-NR, one random split each, trained on LIVE. For both spatial
pooling methods, the loss shows the typical behavior for iterative gradient descent minimization.
Interestingly, WaDIQaM-NR achieves a lower loss than DIQaM-NR during training, but is less capable
to maintain this level of loss during validation and testing.

\subsection{Feature Fusion}
\label{ssec:feature_fusion}
\begin{table}[htb!]
\centering
\caption{LCC Comparison on Different Feature Fusion Schemes}
    \begin{tabular}{ l | l | c | c | c }

    \multirow{2}{*}{Dataset}& \multirow{2}{*}{Method}& \multirow{2}{*}{$f_d -
    f_r$} & concat & concat \\
          &                 &             & ($f_r, f_d$) & ($f_r, f_d, f_d -
          f_r$)\\
     \hline
  \multirow{2}{*}{LIVE}&DIQaM-FR & 0.976 & 0.974 & 0.976 \\
  &WaDIQaM-FR & 0.982  & 0.977 & 0.982\\
 \hline
  \multirow{2}{*}{TID2013}&DIQaM-FR & 0.908 & 0.893 & 0.908 \\
  &WaDIQaM-FR & 0.962  & 0.958 & 0.965\\

    \end{tabular}
\label{tab:feature_fusion}
\end{table}
Results presented for the \ac{FR} models in the previous subsections were obtained employing 
$\textrm{\texttt{concat}}(f_r , f_d, f_d-f_r )$ as feature fusion scheme.
The performances of the three feature fusion schemes are reported for LIVE and
TID2013 in \Tabref{tab:feature_fusion}. Mere concatenation of both feature vectors does not fail but
consistently performs worse than both of the fusion schemes exploiting the explicit
difference of both feature vectors.
This suggests that while the model is able to
learn the relation between the two feature vectors, providing  that relation
explicitly helps to improve the performance.
 The results do not provide enough
evidence for preferring one over the other feature fusion methods. This
might suggest that adding the original feature vectors explicitly to the  representation
does not add  useful information.
Note that the feature fusion scheme might affect the learned features as well, thus, other things
equal, it is not guaranteed the extracted features $f_r$ and $f_d$ are equal for different
fusion methods.

\subsection{Network Depth}
Comparing the proposed \ac{NR} IQA approach to \cite{Kang2014} 
(see \Tabref{tab:performance_comparison}) suggests that the performance of a neural
network-based \ac{IQM} can be increased by making the network
deeper.
In order to evaluate this observation in a \ac{FR} context as well, the
architecture of the WaDIQaM-FR network was modified by removing several layers and
by reducing the intermediate feature vector dimensionality from 512 to 256. This
amounts to the  architecture conv3-32, conv3-32, maxpool, conv3-64, maxpool, conv3-128, maxpool, conv3-256,
maxpool, FC-256, FC-1 with in total ${\raise.17ex\hbox{$\scriptstyle\sim$}}
790\textrm{k}$ parameters instead of ${\raise.17ex\hbox{$\scriptstyle\sim$}}
5.2\textrm{M}$ parameters in the original architecture.
When tested on the LIVE database, the smaller model achieves
a linear correlation of 0.980, whereas the original architecture achieves 0.984. The
same experiment on TID2013 shows a similar result as the shallow model obtains
a linear correlation of 0.949, compared to 0.953 obtained by the deeper model. To
test whether the decrease in model complexity leads to less overfitting and better generalization, the
models trained on TID2013 are additionally tested on CSIQ. The smaller model achieves
a SROCC of 0.911, and is outperformed by the original architecture with a SROCC of 0.927.
The differences are rather small, but it shows that the deeper and more complex
model does lead to a more accurate prediction. However, when computational efficiency is
important, small improvements might not justify the five-fold increase in the number
of parameters.

\subsection{Bridging from Full- to No-Reference Image Quality Assessment}
\begin{figure}[t]
\centering
\subfloat{
\includegraphics[width=0.23\textwidth]{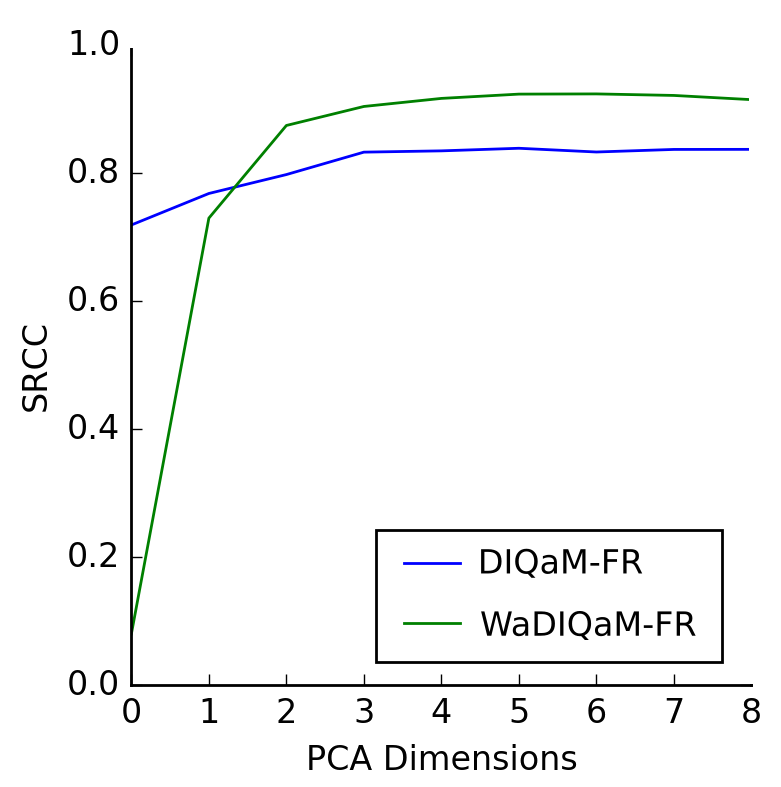}
}
\caption{Average performance on  TID2013 for patchwise dimensionality reduced (Wa)DIQaM-FR 
 in terms of SROCC in dependence of the number of principal components of the reference patch feature vector.
($N_p = 32$)}
\label{fig:rr_tid}
\end{figure}

If argued strictly, the (Wa)DIQaM-Fr as used here is not a \ac{FR}, but a \ac{RR} \ac{IQM}, as only
$N_p=32$ $32\times 32$ patches but not the full image is used for \ac{IQA}. As shown in
\Figref{fig:fr_numberOfPatchesVsPerformance}, reference information could be reduced even further by
reducing the number of patches $N_p$ considered. This can be done without re-training the model.
In the proposed architecture the feature vector extracted from one   
reference patch is 512-dimensional. 

The information available from the reference patch can not only be controlled by steering $N_p$, but also by reducing the dimensionality of the extracted feature vector.
A straight forward approach to do
so would be to systematically change the network architecture from (Wa)DIQaM-FR to (Wa)DIQaM-NR by
reducing the dimensionality of the number of neurons in the last layer of reference branch of the
feature extraction module.
However, this approach would result in a magnitude of models, each trained
for a specific patchwise feature dimensionality.

Another approach is to start with a trained \ac{FR} model and to linearly
reduce the dimensionality of the reference patch feature vector $f_r$  using
\ac{PCA}. 
That way, a network trained for \ac{FR} \ac{IQA} could be used as a
\ac{NR} \ac{IQA} method. In this extreme case the reference feature vector $f_r$ is reduced
to the mean of the reference feature vectors observed in the training data.

PCA is estimated based on the feature vectors
of 4000 reference  patches sampled from the training set and used for
patchwise dimensionality reduction of $f_r$ during testing.
\Figref{fig:rr_tid} shows the performance of this \ac{RR} \ac{IQM} on one
TID2013 test split for increasing dimensionality of the reference
patch feature vectors.
While the dimensionality reduced version of DIQaM-FR is still able to make useful predictions even
without any reference information, this is not the case for the dimensionality reduced version of
WaDIQaM-FR method. This corroborates the previous conjecture that weighted average patch aggregation,
i.e. the reliable estimation of the weights, is more depending on information
from the reference image, at least for homogeneous distortions. 
However, for both DIQaM-FR and WaDIQaM-FR 
3 principal components (dimensions) are already enough to recover the approximate performance  obtained with the
512-dimensional original feature vector. Note that this is the patchwise, not the
imagewise feature dimensionality. 

Although there are studies analyzing the influence of the
feature dimensionality on the performance of \ac{RR} \ac{IQM} systematically
\cite{Soundararajan2012}, we are not aware of any unified framework to study  \ac{NR} to
\ac{FR} approaches.
However, albeit being a promising framework, a fair comparison, e.g. to
the RRED indices \cite{Soundararajan2012} would require an analysis of the  interaction between
patchwise feature dimensionality and number of patches $N_p$ considered.

\section{Discussion \& Conclusion}
\label{sec:discussion}
This paper presented a neural network-based approach to \ac{FR} and \ac{NR}
\ac{IQA} that allows for feature learning and regression in an end-to-end framework. For this, novel
network architectures were presented, incorporating an optional joint
optimization of weighted average patch aggregation implementing a method for 
pooling local patch qualities to global image quality.
To allow for \ac{FR} \ac{IQA},
different feature fusion strategies were studied.
The experimental results show that the
proposed methods outperforms other state-of-the-art approaches for \ac{NR} as
well as for \ac{FR}  \ac{IQA} and achieve generalization capabilities
competitive to state-of-the-art data-driven  approaches.
However, as for all current
data-driven \ac{IQA} methods generalization performance offers considerable room for improvement.
A principle problem for data-driven \ac{IQA} is the
relative lack of data and significantly larger databases are hopefully to be
expected any time soon, potentially using new methods of psychophysiological
assessment \cite{Bosse2017,bosse2016smc,Engelke2016}.
Until then, following the BIECON \cite{kim2017fully} framework,
networks could be pre-trained unsupervised by optimizing on reproducing the quality prediction of
a \ac{FR} \ac{IQM}, and a pre-trained network employing patch-wise weighting could be refined by
end-to-end training. This combines the advantages of the two
approaches.

Even though a relative generic neural network is able to achieve high prediction performance,
incorporating \ac{IQA} specific adaptations to the architecture may lead
to further improvements. Our results show that there is room for optimization
in terms of feature dimensionality and balancing the ratio between network
parameters. Here, prediction performance and generalization ability is important to
be studied.
In this work, we optimized based on \ac{MAE}. However, \ac{IQA} methods are
commonly evaluated in terms of correlation and other loss function might be more feasible.
We sketched how the proposed framework could be used for \ac{RR} \ac{IQA}. We
did not present a full-fledged solution, but believe that the results indicate an
interesting starting point.
 
Local weighting of distortion is not a new concept.
Classical approaches usually compute the local weights based  on a saliency
model from the reference image \cite{Zhang2016}, or the reference and the
distorted image \cite{Zhang2014}.
Selecting an appropriate saliency model is crucial for the success of this
strategy and models best at predicting saliency are not necessarily best for
\ac{IQA} \cite{Zhang2016}.
The proposed weighted average patch aggregation method allows for a joint,
end-to-end optimization of local quality estimation and weight assignment.

The equivalence of our weighting scheme to Eq. 1 allows us to interpret the two learned maps
as a weighting map and a quality map.  Thus, the formal equivalence with the classical approach of
linear distortion weighting suggests that local weights capture local saliency. However, this is not
necessarily true, as the optimization criterion is not saliency, but image quality. In fact, we
showed that the local weights not only depend on the structural (and potentially semantical)
organization of the reference image, but also on the distortion type and the spatial distribution of
the distortion. This is a fundamental problem for IQA and future work should address the conceptual
similarity between the learned weights and saliency models in greater detail.       

The proposed network could be adapted
for end-to-end learning local weights only in order to improve the prediction
performance of any given \ac{IQM}. This could be combined with conventional regression \cite{BosICIP17}.
Also explanation methods \cite{BachPLOS15, MonArXiv17} can be applied to better understand what features were actually learned by the network.
From  an application-oriented perspective the proposed method may be adapted and
evaluated for quality assessment of 3D images and 2D and 3D videos.

Although there are still many obstacles and challenges for purely data-driven
approaches, the performance of the presented approach and, given the fact that no  domain
knowledge is necessary, its relative simplicity suggests that neural networks used for \ac{IQA} have
lots of potential and are here with us to stay.

\FloatBarrier

%
\bibliographystyle{myIEEEtran}
%
\bibliography{IEEEabrv,2016_journal_cnnIqa_new,2016_journal_cnnIqa3}
\end{document}